\documentclass{article} 
\usepackage{iclr2026_conference,times}

\usepackage{amsmath,amsfonts,bm}

\def\eqref#1{equation~\ref{#1}}

\def\1{\bm{1}}

\DeclareMathAlphabet{\mathsfit}{\encodingdefault}{\sfdefault}{m}{sl}
\SetMathAlphabet{\mathsfit}{bold}{\encodingdefault}{\sfdefault}{bx}{n}

\usepackage{afterpage}
\usepackage{hyperref}
\usepackage{url}
\usepackage{amsthm}
\usepackage{algorithm}
\usepackage{enumitem}
\usepackage{algorithmic}
\usepackage{wrapfig}
\usepackage{capt-of}
\usepackage{booktabs}
\usepackage{float}
\usepackage{comment}
\usepackage{subcaption}
\usepackage{setspace} 
\usepackage{multirow}
\usepackage{mathtools}
\usepackage[capitalize,nameinlink]{cleveref}
\usepackage[most]{tcolorbox}
\usepackage{bbm}
\usepackage{pifont} 
\usepackage{icomma}
\usepackage[%
    prefixes-as-symbols=false,
    ]{siunitx}
\usepackage{tablefootnote}
\usepackage{adjustbox}
\usepackage[most]{tcolorbox} 
\tcbuselibrary{skins} 
\usepackage{fontawesome5} 
\usepackage{xcolor} 
\usepackage{makecell} 
\usepackage[table]{xcolor} 
\usepackage{colortbl}      
\usepackage{hyperref}
\usepackage{romannum}

\definecolor{Gray}{gray}{0.9}
\definecolor{Red}{rgb}{0.768, 0.054, 0.054}
\definecolor{Blue}{rgb}{0.152, 0.294, 0.925}
\definecolor{URL}{rgb}{0,0.4,0.7}
\hypersetup{
    colorlinks=true,
    citecolor=teal,
    linkcolor=Red,
    urlcolor=URL,
    }
\definecolor{Green}{rgb}{0,0.9,0}

\usepackage{pifont}
\newcommand{\xmark}{\ding{55}}

\tcbset{
    llmprompt/.style={
        colback=gray!10, 
        colframe=blue!50, 
        fonttitle=\bfseries, 
        title=Prompt Format, 
        sharp corners, 
        boxrule=1pt, 
        width=\textwidth, 
        before=\vspace{0.3\baselineskip}, 
        after=\vspace{0.3\baselineskip}, 
    }
}

\AtBeginDocument{\pagenumbering{arabic}}

\DeclarePairedDelimiterX{\infdivx}[2]{(}{)}{%
  #1\;\delimsize\|\;#2%
}

\newcommand{\ie}{\textit{i.e.}}

\newcommand{\eg}{\textit{e.g.}}

\Crefname{table}{Tab.}{Tabs.}
\Crefname{figure}{Fig.}{Figs.}
\Crefname{equation}{Eq.}{Eqs.}
\Crefformat{section}{#2\S#1#3}

\title{Automated Structured Radiology Report Generation with Rich Clinical Context}

\author{%
    Seongjae Kang$^{\spadesuit,\dagger}$ \quad
    Dong Bok Lee$^{\clubsuit,\dagger}$ \quad
    Juho Jung$^{\spadesuit}$ \quad
    \textbf{Dongseop Kim}$^{\spadesuit}$ \\
    \textbf{Won Hwa Kim}$^{\diamondsuit}$ \quad
    \textbf{Sunghoon Joo}$^{\spadesuit}$ \\[0.1em]
    $^{\spadesuit}$VUNO Inc. \quad
    $^{\clubsuit}$KAIST \quad
    $^{\diamondsuit}$POSTECH \\[0.1em]
    \texttt{\{seongjae.kang, juho.jung, dongseop.kim, sunghoon.joo\}@vuno.co} \\
    \texttt{markhi@kaist.ac.kr, wonhwa@postech.ac.kr} \\[0.1em]
    \footnotesize{$^{\dagger}$Equal contribution}
}

\iclrfinalcopy 
\begin{document}

\maketitle
\vspace{-0.1in}
\definecolor{linkcolor}{RGB}{5, 70, 170}

\begin{abstract}
Automated \emph{structured radiology report generation} (SRRG) from chest X-ray images offers significant potential to reduce workload of radiologists by generating reports in structured formats that ensure clarity, consistency, and adherence to clinical reporting standards.
While radiologists effectively utilize available clinical contexts in their diagnostic reasoning, existing SRRG systems \emph{overlook} these essential elements.
This fundamental gap leads to critical problems including \emph{temporal hallucinations} when referencing non-existent clinical contexts.
To address these limitations, we propose \emph{contextualized SRRG} (\textbf{C-SRRG}) that comprehensively incorporates rich clinical context for SRRG.
We curate C-SRRG dataset by integrating comprehensive clinical context encompassing 1) multi-view X-ray images, 2) clinical indication, 3) imaging techniques, and 4) prior studies with corresponding comparisons based on patient histories.
Through extensive benchmarking with state-of-the-art multimodal large language models, we demonstrate that incorporating clinical context with the proposed C-SRRG significantly improves report generation quality, as summarized in \Cref{fig:main_performance}.
We publicly release dataset, code, and checkpoints to facilitate future research for clinically-aligned automated RRG at \href{https://github.com/vuno/contextualized-srrg}{\textcolor{linkcolor}{https://github.com/vuno/contextualized-srrg}}.
\end{abstract}
\begin{figure}[h]
\vspace{-0.1in}
    \centering
    \includegraphics[width=0.82\textwidth]{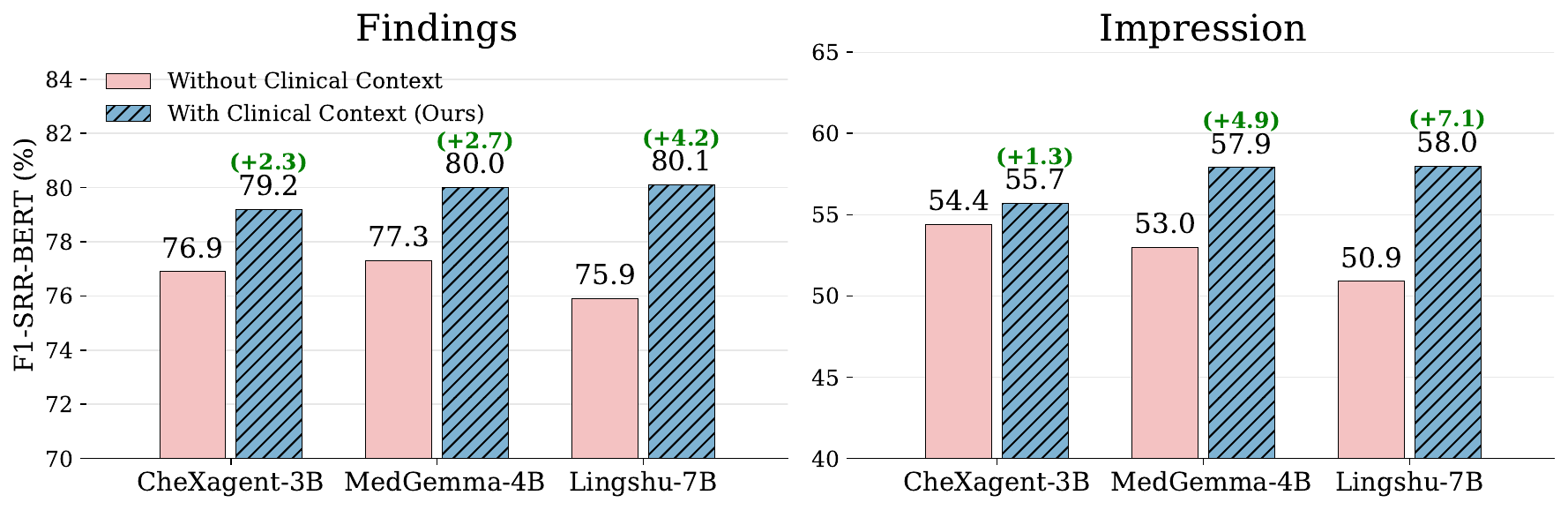}
    \vspace{-0.1in}
    \caption{\textbf{Clinical context consistently and significantly improves medical MLLMs}—including CheXagent-3B~\citep{chen2024chexagent}, MedGemma-4B~\citep{sellergren2025medgemma}, and Lingshu-7B~\citep{xu2025lingshu}—on both the findings and impression tasks for SRRG, as measured by F1-SRR-BERT metric~\citep{delbrouck2025automated}.
    Clinical context becomes \textbf{increasingly critical as MLLMs scale up}, highlighting its importance in RRG.}
    \label{fig:main_performance}
\vspace{-0.1in}
\end{figure}

\section{Introduction}
\vspace{-5pt}


Writing a radiology report requires radiologists to accurately interpret images and synthesize them into two main components: 1) detailed \emph{findings} that systematically document anatomical structures and pathological observations, and 2) concise \emph{impressions} that provide clinical interpretations for subsequent decision-making~\citep{wallis2011radiology, pahadia2020radiology, haygood2018consultation, trinh2019radiology, european2011good}.
However, generating such comprehensive reports is both cognitively demanding and time-consuming for radiologists.
Given the high volume of imaging studies and the time-intensive nature of report writing, there is a critical need for automated systems that can assist radiologists by generating accurate, structured reports while reducing radiologists' workload and improving diagnostic efficiency~\citep{markotic2021radiologist,alexander2022mandating}.


\emph{Automated radiology report generation} (\textbf{RRG}) has emerged as a crucial task to address these challenges by assisting radiologists in the diagnostic workflow~\citep{esteva2019guide,sloan2024automated,tanno2024collaboration}.
Deep learning has accelerated the development of automated RRG frameworks that generate reports directly from medical images~\citep{shin2016learning,jing2018automatic,li2018hybrid,wang2018tienet,jing2019show,chen2020memorydriven,chen2022cross,wang2022cross}.
Recent advances in multimodal large language models (MLLMs) further enhanced this capability by integrating vision foundation models with large language models capable of generating coherent and clinically relevant text~\citep{lee2023cxrllava,li2023llava,hyland2023maira,bannur2024maira2,chen2024chexagent,sellergren2025medgemma,xu2025lingshu,wang2025capabilities}.

\begin{figure}[t]
    \vspace{-0.1in}
    \centering
    \includegraphics[width=0.98\textwidth]{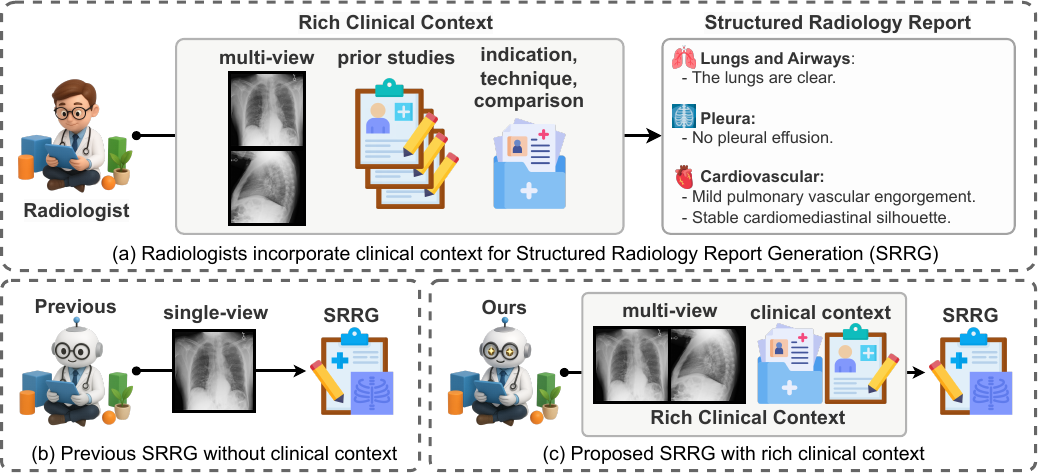}
    \vspace{-0.1in}
    \caption{\textbf{A conceptual illustration of the proposed C-SRRG}. \textbf{(a)} Radiologists routinely use clinical context, while \textbf{(b)} existing SRRG frameworks do not. Motivated by this gap, \textbf{(c)} C-SRRG leverages multi-view images, indication, technique, and variable-length prior studies/comparisons to generate structured radiology reports.}
    \label{fig:concept}
    \vspace{-0.24in}
\end{figure}


Despite the progress, most automated RRG frameworks overlook \emph{essential clinical context} such as imaging indication, technique, and prior studies that radiologists use to generate reports~\citep{kahn2009toward,european2011good}.
Ignoring the clinical context leads to systematic errors~\citep{liu2019clinically,ramesh2022improving} as the models fail to capture patient-specific properties and longitudinal changes essential for accurate diagnosis, including \emph{temporal hallucinations} where the models generate references to nonexistent priors or fabricate temporal comparisons (\Cref{fig:hallucination_example1,fig:hallucination_example2,fig:hallucination_example3}).
Although some work injected partial context---\eg, multi-view images~\citep{yuan2019automatic,chen2022cross}, historical images~\citep{hou2023recap,zhu2023longitudinal}, and prior reports or indications~\citep{chen2024evoke,wang2024hergen}---these are still \emph{limited}, \eg, only consider partial clinical context, rely only on the immediately preceding image or report~\citep{bannur2024maira2,liu2025enhanced,liu2025priorrg}, with all approaches targeting free-form RRG rather than \emph{structured} RRG (SRRG)~\citep{delbrouck2025automated}.

To this end, we first present \textbf{C-SRRG}, a framework for \emph{contextualized structured radiology report generation} (\Cref{fig:concept}), built upon the recently introduced SRRG paradigm~\citep{delbrouck2025automated}.
We curate a \emph{large-scale} dataset for \emph{structured} report generation with \emph{rich clinical context}, by leveraging MIMIC-CXR~\citep{johnson2019mimiccxr} and CheXpert Plus~\citep{chambon2024chexpert}.
Specifically, our \textbf{C-SRRG dataset} provides 1) \textbf{multi-view images} (frontal and lateral), 2) clinical \textbf{indication}, 3) imaging \textbf{technique}, and 4) variable-length \textbf{prior studies} with corresponding comparisons, which models can incorporate depending on their architecture.


We evaluate the effectiveness of the proposed C-SRRG with state-of-the-art (SoTA) medical MLLMs---including CheXagent-3B~\citep{chen2024chexagent}, MedGemma-4B~\citep{sellergren2025medgemma}, and Lingshu-7B~\citep{xu2025lingshu}---and find that incorporating clinical context \textbf{substantially and consistently improves report quality} (summarized in \Cref{fig:main_performance} and detailed in \Cref{tab:findings_results,tab:impression_results}) measured by various metrics~\citep{papineni2002bleu, lin2004rouge, zhang2019bertscore, delbrouck2022improving, delbrouck2025automated}. Interestingly, the clinical context \textbf{becomes increasingly critical as the models scale up} from 3B to 7B.
We also provide a comprehensive analysis, including extensive ablation studies on clinical context (\Cref{tab:findings_results_context,tab:impression_results_context,tab:ablation_findings,tab:ablation_impression}), temporal hallucination mitigation (\Cref{tab:hallucination_results}), and organ-level performance (\Cref{tab:organ_level_comparison_validation}). We will \emph{publicly release} the 1) \textbf{dataset}, 2) \textbf{code}, and 3) \textbf{checkpoints} of benchmarked models to facilitate further research in C-SRRG and benefit the community.

Our contributions and empirical findings are summarized as follows:
\vspace{-0.05in}
\begin{itemize}[itemsep=1mm,parsep=1pt,topsep=0pt,leftmargin=*]



\item We identify a key limitation of existing SRRG frameworks, \ie, the neglect of essential \emph{clinical context}, which induces systematic errors, most notably \emph{temporal hallucinations} about nonexistent prior studies. To address this, we introduce a clinically contextualized SRRG framework (\textbf{C\text{-}SRRG}) that explicitly integrates clinical context into the generation process.


\item We curate \emph{the largest structured} radiology report generation dataset with \emph{rich clinical context}, namely, \textbf{C-SRRG dataset}, which includes 1) multi-view images, 2) indication, 3) technique, and 4) prior studies/comparisons, for training and evaluation of the proposed C-SRRG framework.




\item As summarized in \Cref{fig:main_performance}, we provide a \emph{comprehensive benchmark} of \textbf{SoTA MLLM}-based SRRG models, demonstrating that clinical context becomes \textbf{increasingly critical as models scale up}---enhancing report quality (\eg, +2.3$\sim$4.2/+1.3$\sim$7.1 on findings/impression for F1-SRR-BERT) while reducing \emph{temporal hallucinations} (\Cref{tab:hallucination_results}; \eg, 12.2\%/18.0\% on findings/impression).


\vspace{-0.05in}
\end{itemize}

\vspace{-5pt}
\section{Related Work}
\vspace{-10pt}

\paragraph{Automated radiology report generation (RRG).}
Automated RRG has emerged as a promising approach to reduce radiologists' workload and improve reporting efficiency~\citep{yang2023radiology, sloan2024automated, esteva2019guide, sirshar2022attention, tanno2024collaboration, chestx2025transcribe}.
While early approaches simply combined vision encoders with language decoders for visual feature extraction~\citep{he2016deep, dosovitskiy2021an} and text generation~\citep{shin2016learning, jing2018automatic, li2018hybrid, wang2018tienet, jing2019show, chen2020generating, yan2022clinical, miura2021improving}, architectural innovations have significantly improved report quality, such as memory-driven transformers~\citep{chen2020memorydriven, liu2024factual}, specialized architectures for medical domain knowledge~\citep{yang2021radiology,wang2022medical, kong2022transq}, cross-modal learning for improved alignment~\citep{chen2022cross, wang2022cross, li2023dynamic}, and region-guided frameworks for anatomically relevant features~\citep{tanida2023interactive, li2023unar, hou2023organ}. 
In this work, we focus on extending the recently proposed structured RRG \citep[SRRG;][]{delbrouck2025automated}---improving clarity, consistency, and interpretability through standardized structure~\citep{weiss2008structured,kahn2009toward,bosmans2012structured,bosmans2015structured}---by incorporating rich clinical context aligned with radiologists' workflow.

\vspace{-0.1in}
\paragraph{Multimodal large language models (MLLMs).}
Building on recent advances in LLMs~\citep{bai2023qwen,achiam2023gpt,touvron2023llama,touvron2023llama2,yang2025qwen3}, MLLMs have shown strong performance across many domains, including medical applications~\citep{achiam2023gpt,team2023gemini,yang2025qwen3,comanici2025gemini,wang2025capabilities,zhu2025internvl3}.
They integrate visual understanding with natural-language generation, enabling effective tools for medical image analysis and clinical text generation~\citep{li2023llava,chen2024huatuogpt,he2024gsco,hurst2024gpt,lai2025med,pan2025medvlm}.
Medical-specific MLLMs further improve performance by incorporating domain knowledge and clinical expertise through specialized training procedures, including CheXagent~\citep{chen2024chexagent}, MedGemma~\citep{sellergren2025medgemma}, and Lingshu~\citep{xu2025lingshu}.
These foundation models are particularly promising for comprehensive RRG frameworks~\citep{lee2023llm,zhu2023minigpt,liu2024icl,wang2023r2gengpt,hyland2023maira,bannur2024maira2,lee2023cxrllava,chen2024chexagent,dai2025qoq,sellergren2025medgemma,xu2025lingshu}, where their ability to accept flexible inputs and produce coherent clinical text is especially valuable.
Accordingly, we benchmark medical MLLMs for contextualized SRRG.

\vspace{-0.1in}
\paragraph{Clinical context.}
Radiologists routinely leverage clinical context when drafting reports, drawing upon patient history, prior studies, and clinical indications~\citep{kahn2009toward,european2011good,wallis2011radiology,haygood2018consultation,trinh2019radiology,pahadia2020radiology,castillo2021effect,nguyen2021automated}, motivating various approaches to integrate such clinical context into automated RRG frameworks.
Multi-view image analysis utilizes complementary imaging perspectives, such as frontal and lateral views, to provide comprehensive anatomical coverage~\citep{yuan2019automatic, chen2024evoke, chen2022cross, nooralahzadeh2021progressive, serra2023controllable, wang2023observation, van2023longitudinal}.
Indication and clinical history integration approaches incorporate patient-specific clinical information~\citep{hou2023recap, zhu2023longitudinal, wang2024hergen, zhang2023structural, chen2024evoke}.
Previous studies enable temporal comparison and disease progression tracking~\citep{hou2023recap, zhu2023longitudinal, serra2023controllable, wang2024hergen, liu2021exploring}.
Recent works such as MLRG~\citep{liu2025enhanced}, PriorRG~\citep{liu2025priorrg}, and MAIRA-2~\citep{bannur2024maira2} have attempted to incorporate clinical context for more comprehensive report generation.
However, these approaches have limitations: they either 1) consider only partial clinical context, 2) are restricted to specific input configurations, 3) have narrow temporal scope (only the previous prior study), with all 4) focusing on unstructured free-form report generation.


\vspace{-5pt}
\section{Method}
\label{sec:method}
\vspace{-10pt}

In this section, we first elaborate on the dataset curation process for contextualized radiology report generation (C-SRRG) in \Cref{sec:dataset} and then detail the proposed C-SRRG framework in \Cref{sec:c-srrg}.

\vspace{-0.05in}
\subsection{Curation of Contextualized Clinical Context}
\label{sec:dataset}
\vspace{-5pt}

\begin{figure}[t]
\begin{tcolorbox}[
    colback=red!5,
    colframe=red!40,
    width=\textwidth,
    boxrule=0.8pt,
    arc=2mm,
    left=3mm,
    right=3mm,
    top=2mm,
    bottom=2mm
]
\small
\textbf{Structured Report (Excerpt):}\\
History: A male patient with hep C cirrhosis and large right pleural effusion status post thoracocentesis.
\\[0.3em]
Comparison: Prior portable AP chest radiograph
\\[0.3em]
Findings:
\\[0.3em]
Pleura:\\
- Moderate pleural effusion within the right pleural space.\\
- Moderate right pneumothorax, \colorbox{red!20}{\textbf{new from prior exam}.}\\
- No left pleural effusion or pneumothorax.
\\[0.3em]
Impression:\\
1. Moderate right-sided pneumothorax.\\
2. Moderate right pleural effusion.
\begin{tcolorbox}[
    colback=yellow!10,
    colframe=orange!60,
    boxrule=1pt,
    arc=1mm,
    left=2mm,
    right=2mm,
    top=1mm,
    bottom=1mm
]
\textbf{Hallucination:} The phrase ``\textbf{new from prior exam}'' represents temporal information that cannot be verified from the current study alone, if not with previous history.
\end{tcolorbox}
\end{tcolorbox}
\vspace{-0.1in}
\caption{\textbf{An example of temporal hallucinations}.
This report contains ``\textbf{new from prior exam}'' even though any prior studies are not provided.
Please see examples of full structured reports in \Cref{fig:hallucination_example1,fig:hallucination_example2,fig:hallucination_example3}.}
\label{fig:hallucination_example1_excerpt}
\vspace{-0.2in}
\end{figure}

\paragraph{Motivation.}
Our design principle is to reflect the clinical workflow of radiologists that incorporates a diverse diagnostic context such as indication, technique, and comparison~\citep{wallis2011radiology, trinh2019radiology, pahadia2020radiology, nguyen2021automated}, supported by empirical evidence showing improvement in report quality~\citep{castillo2021effect,liu2021exploring}.
This emphasis on a comprehensive clinical context aligns with recent work advocating that AI systems must move beyond narrow task-specific approaches that lack the ability to incorporate multimodal data and provide comprehensive interpretation assistance~\citep{Dogra2025}.
Most importantly, without this context, existing automated systems are prone to \emph{temporal hallucinations}: ground truth reports frequently contain temporal statements such as ``\textbf{new from prior exam}'' (as shown in \Cref{fig:hallucination_example1_excerpt}), which leads models to hallucinate by referencing nonexistent prior examination~\citep{ramesh2022improving}.
When trained on such data, the SRRG frameworks learn to generate these temporal phrases even when no prior studies are available, as demonstrated in \Cref{fig:trained_hallucination_example1,fig:trained_hallucination_example2,fig:trained_hallucination_example3,fig:trained_hallucination_example4,fig:trained_hallucination_example5,fig:trained_hallucination_example6}.

\vspace{-0.1in}
\paragraph{Clinical context.}
To address this limitation, we incorporate rich clinical context into automatic SRRG frameworks.
Specifically, we consider \textbf{four clinical elements} that radiologists routinely use:

\begin{enumerate}[itemsep=1mm,parsep=1pt,topsep=0pt,leftmargin=*]
\vspace{-0.05in}
\item \textbf{Multi-view images} (\eg, posteroanterior, anteroposterior, and lateral) provide complementary perspectives from different angles, enabling comprehensive assessment and detection of abnormalities that may be obscured in single views.
Multi-view fusion captures richer information through cross-view consistency, improves pathological localization accuracy, and reduces diagnostic uncertainty~\citep{yuan2019automatic,chen2024evoke}.

\item \textbf{Indication} conveys the clinical rationale for imaging, providing context about patient symptoms, suspected conditions, or clinical questions.
This enables models to focus on specific diagnostic questions, tailor findings to physician concerns, and avoid clinically insignificant findings.


\item \textbf{Technique} documents examination parameters and limitations including imaging protocols, contrast use, and factors affecting image quality. It helps models note technical caveats, avoid mistaking artifacts for pathology, and prevent duplicate exams.

\item \textbf{Prior studies}, when available, enable temporal \textbf{comparison} by providing a history to detect disease progression, treatment response, and interval changes.
Radiologists routinely consult such a history~\citep{haygood2018consultation,liu2025priorrg}, which supports accurate change detection and prevents hallucinations referencing nonexistent prior exams.
\vspace{-0.05in}
\end{enumerate}

\begin{table}[t]
    \centering    
    \vspace{-0.1in}
    \caption{\textbf{Dataset statistics} for C-SRRG-Findings and C-SRRG-Impression.}
    \vspace{-0.1in}
    \label{tab:dataset_statistics}
    \resizebox{0.7\textwidth}{!}{%
        \begin{tabular}{@{}lccccc@{}}
        \toprule
        \textbf{Tasks} & \textbf{Train} & \textbf{Valid} & \textbf{Test} & \textbf{Test-reviewed} & \textbf{Total} \\
        \midrule
        Findings & 181,874 & 976 & 1,459 & 233 & 184,542 \\
        Impression & 405,972 & 1,505 & 2,219 & 231 & 409,927 \\
        \bottomrule
        \end{tabular}%
    }
    \vspace{-0.1in}
\end{table}

\begin{figure}[H]
    \vspace{-0.1in}
    \centering
    \includegraphics[width=0.98\textwidth]{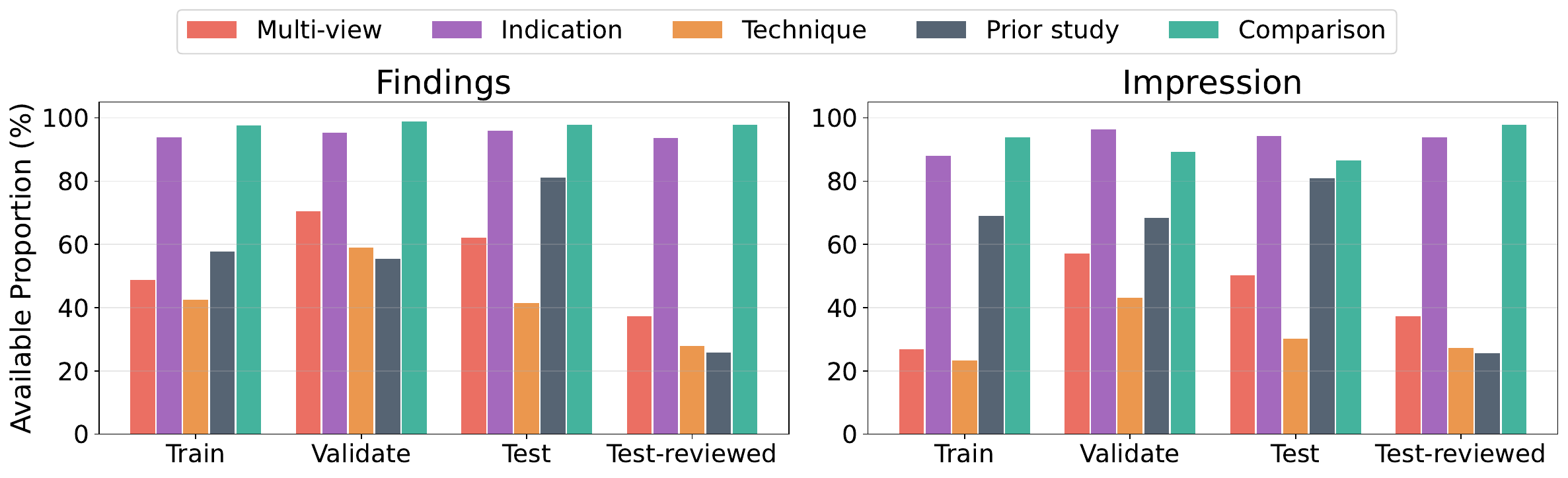}
    \vspace{-0.1in}
    \caption{\textbf{Available proportion of clinical context} for each split in findings and impression.}
    \vspace{-0.1in}
    \label{fig:available_clinical_context}
\end{figure}

\begin{figure}[H]
    \vspace{-0.15in}
    \centering
    \includegraphics[width=0.98\textwidth]{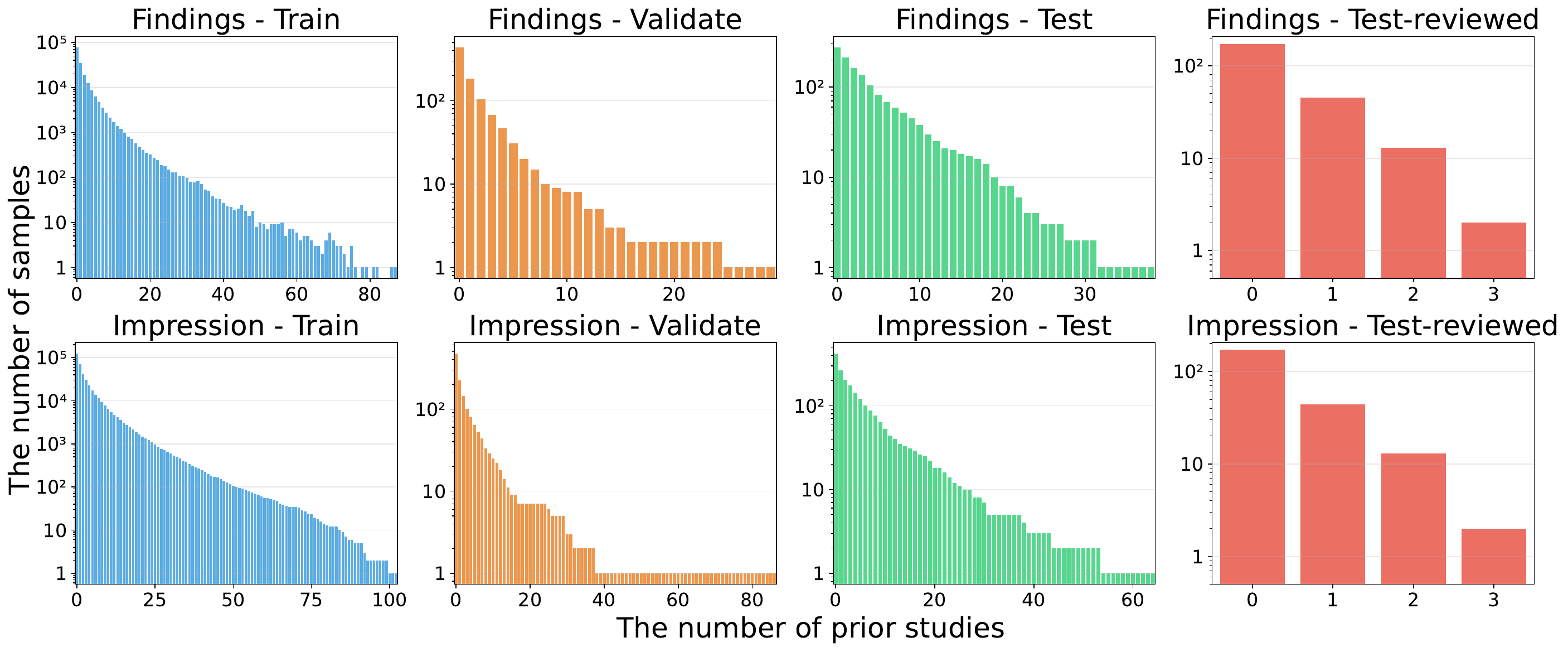}
    \vspace{-0.1in}
    \caption{\textbf{Distribution of the number of prior studies} available per sample for findings and impression.}
    \label{fig:previous_history_count}
    \vspace{-0.15in}
\end{figure}

\vspace{-0.1in}
\paragraph{Dataset curation.}

We build on the recently proposed SRRG dataset~\citep{delbrouck2025automated}, which includes both MIMIC~\citep{johnson2019mimiccxr} and CheXpert Plus~\citep{chambon2024chexpert}.
We employ dataset-specific approaches to extract the necessary clinical context.
When multiple views are available, we integrate multi-view images using \texttt{ViewPosition} for MIMIC and \texttt{frontal\_lateral}, \texttt{ap\_pa} for CheXpert Plus.
For MIMIC, each patient is identified by a unique \texttt{subject\_id} with associated \texttt{StudyDate} and \texttt{StudyTime} fields.
We group patients by \texttt{subject\_id}, then use temporally-ordered \texttt{StudyDate} and \texttt{StudyTime} to establish chronological sequences.
For CheXpert Plus, each study contains a \texttt{deid\_patient\_id} and \texttt{patient\_report\_date\_order} field.
We group studies by patient and use the order of report dates to form longitudinal sequences.
For other clinical contexts (indication, technique, comparison), we use SRRG components, parsed from free-form reports using GPT-4~\citep{achiam2023gpt}.

\begin{wrapfigure}{r}{0.56\textwidth}
    \centering
    \vspace{-0.2in}    \includegraphics[width=0.56\textwidth]{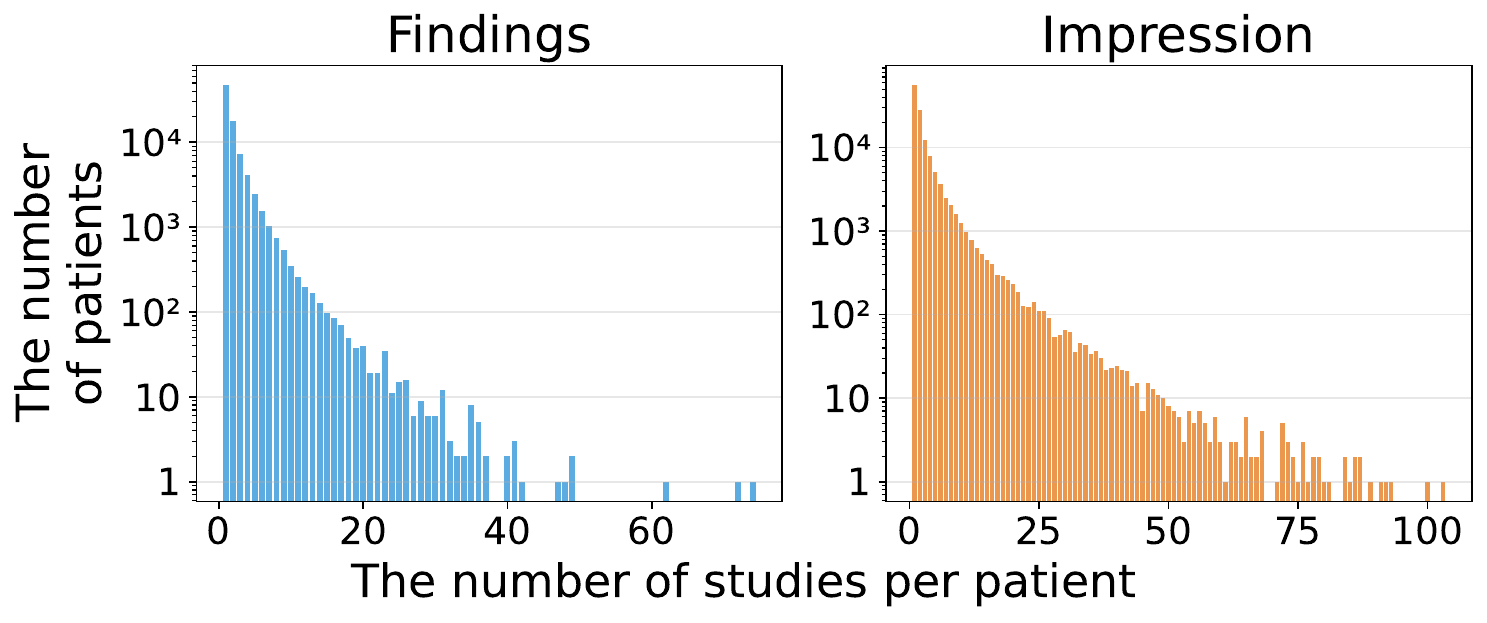}
    \vspace{-0.25in} 
    \caption{\textbf{Distribution of the number of studies per patient}.}
    \vspace{-0.15in}
    \label{fig:patient_study_count_comparison}
\end{wrapfigure}
\vspace{-0.1in}

\paragraph{Dataset analysis.} 
Accordingly, we curate two C-SRRG tasks that mirror clinical practice—\textbf{C-SRRG-Findings} and \textbf{C-SRRG-Impression} with \textbf{train}, \textbf{valid}, \textbf{test}, and \textbf{test-reviewed}\footnote{The \emph{test-reviewed} split reports are reviewed by board-certified radiologists~\citep{delbrouck2025automated}.} splits, as summarized in \Cref{tab:dataset_statistics}.
The splits enforce the strict separation of patients between training and evaluation to prevent data leakage and properly assess generalization, as in \Cref{fig:patient_overlap_heatmaps}.
The availability of a clinical context varies across splits and tasks (\Cref{fig:available_clinical_context}).
The availability of prior studies follows a long-tailed distribution (\Cref{fig:previous_history_count}), alongside the long-tailed counts of studies per patient (\Cref{fig:patient_study_count_comparison})---from no history to extensive longitudinal sequences. This variability reflects \emph{real-world clinical practice} and requires models to handle missing information while leveraging available context.

\begin{figure}[t]
\centering
\vspace{-0.2in}
\begin{tcolorbox}[
    colback=gray!5,
    colframe=gray!40,
    width=\textwidth,
    boxrule=0.8pt,
    arc=2mm,
    left=3mm,
    right=3mm,
    top=2mm,
    bottom=2mm
]
\small
\textbf{SYSTEM PROMPT:}\\
You are an expert radiologist.
\\[1.5em]
\textbf{USER PROMPT:}\\
Analyze the current chest X-ray images and compare them with the previous studies to write the IMPRESSION section of a radiology report.
Provide a concise clinical summary and diagnosis, noting any changes from the prior studies, focusing on the most recent comparisons.
Consider the available clinical contexts when formulating your impression.
\end{tcolorbox}
\vspace{-0.2in}
\caption{\textbf{Prompt format for IMPRESSION} for C-SRRG.}
\label{fig:prompt_format_impression}
\vspace{-0.02in}
\end{figure}
\begin{figure}[t]
\centering
\vspace{-0.1in}
\begin{tcolorbox}[
    colback=gray!5,
    colframe=gray!40,
    width=\textwidth,
    boxrule=0.8pt,
    arc=2mm,
    left=3mm,
    right=3mm,
    top=2mm,
    bottom=2mm
]
\small
\textbf{=== CURRENT CLINICAL HISTORY/INDICATION ===}\\
Male with end-stage renal disease on hemodialysis, multiple orthopedic hardware, bioprosthetic aortic valve replacement, coronary artery disease status post coronary artery bypass grafting, permanent pacemaker for sick sinus syndrome, admitted for MRSA bacteremia, transferred to the critical care unit for hypotension during anesthesia induction.
Patient intubated for respiratory status assessment.
\\[1.0em]
\textbf{=== CURRENT TECHNIQUE ===}\\
Chest single view
\\[1.0em]
\textbf{=== CURRENT COMPARISON ===}\\
Prior imaging at an unspecified time.
\\[1.0em]
\textbf{=== CURRENT IMAGES ===}\\
\texttt{[Images: Current chest X-ray images]}
\end{tcolorbox}
\vspace{-0.2in}
\caption{\textbf{An example of current study} for C-SRRG.}
\label{fig:current_study_format}
\vspace{-0.1in}
\end{figure}
\begin{figure}[!t]
\centering
\begin{tcolorbox}[
    colback=gray!5,
    colframe=gray!40,
    width=\textwidth,
    boxrule=0.8pt,
    arc=2mm,
    left=3mm,
    right=3mm,
    top=2mm,
    bottom=2mm
]
\small
\textbf{ASSISTANT RESPONSE:}\\
IMPRESSION:\\
1. Worsened left basilar opacity, which may represent atelectasis or pneumonitis in the appropriate clinical context.\\
2. Mild improvement in pulmonary vascularity.
\end{tcolorbox}
\vspace{-0.2in}
\caption{\textbf{An example of ground-truth assistant responses} in the C-SRRG-Impression dataset.}
\label{fig:response_format_impression}
\vspace{-0.22in}
\end{figure}

\vspace{-5pt}
\subsection{Contextualized Radiology Report Generation (C-SRRG)}\label{sec:c-srrg}
\vspace{-5pt}

\paragraph{Prompt design.}
We construct prompt templates for four core settings: 1) findings prompts with prior studies (\Cref{fig:prompt_format_finding}), 2) findings prompts without prior studies (\Cref{fig:prompt_format_finding_no_history}), 3) impression prompts with prior studies (\Cref{fig:prompt_format_impression}), and 4) impression prompts without prior studies (\Cref{fig:prompt_format_impression_no_history}).
Each prompt consists of the clinical context (\ie, indication, technique, and comparison for the current study), and the associated images (\Cref{fig:current_study_format}).
When available, it incorporates prior studies that also include indication, technique, comparison, and reports on findings or impression (\Cref{fig:prior_study_format}).
These structured components are concatenated to form a \emph{single multimodal token sequence}.
As shown in \Cref{fig:response_format_findings,fig:response_format_impression}, the response format is standardized for the generation of structured reports.
Detailed examples of prompt structures and integration of clinical context can be found in \Cref{sec:ift_prompt} and \Cref{sec:ift_prompt_full}.

\vspace{-0.1in}
\paragraph{Training and inference.}
We fine-tune medical MLLMs on these contextualized prompt--response pairs for both findings and impression tasks. Models receive prompt and clinical context to form a unified multimodal input sequence. The training objective is then the next-token prediction task under an autoregressive language modeling loss: $\frac{1}{T}\sum_{t=1}^{T} -\log p_{\theta}(y_t|x, y_{<t}),$
where \(x\) is the multimodal token sequence comprising the prompt (\Cref{fig:prompt_format_finding,fig:prompt_format_finding_no_history,fig:prompt_format_impression,fig:prompt_format_impression_no_history}), the clinical context of the current study (\Cref{fig:current_study_format}) and any prior studies (\Cref{fig:prior_study_format}), and \(y_{1:T}\) is the target token sequence (\eg, reports on findings or impression; \Cref{fig:response_format_findings,fig:response_format_impression}). Here, $p_{\theta}$ denotes the MLLM parameterized by $\theta$. We minimize the negative log-likelihood with respect to $\theta$, \ie, standard cross-entropy over the vocabulary. If prior studies are available, they are inserted into designated slots; otherwise, the model receives only the clinical context of the current study.
This design allows the model to adapt to \emph{heterogeneous clinical contexts} (\Cref{fig:available_clinical_context,fig:previous_history_count,fig:patient_study_count_comparison}), to produce context-aware reports when there is prior information, and to avoid hallucinated temporal comparisons when not.



\vspace{-0.2in}
\section{Experiments}
\label{sec:experiments}
\vspace{-0.06in}


\subsection{Experimental Setups}
\label{sec:experimental_setups}
\vspace{-0.05in}

\begin{wraptable}{r}{0.2\textwidth}
\vspace{-0.2in}
\centering
\caption{\small\textbf{Summary of hyperparameters}.}
\vspace{-0.1in}
\label{tab:hyperparameters}
\resizebox{0.2\textwidth}{!}{%
\begin{tabular}{@{}lc@{}}
\toprule
\textbf{Name} & \textbf{Value} \\
\midrule
\rowcolor{Gray}
\multicolumn{2}{@{}l}{\textit{LoRA}} \\
Rank $r$ & 32 \\
$\alpha$ & 64 \\ 
Dropout $p$ & 0.1 \\
\midrule
\rowcolor{Gray}
\multicolumn{2}{@{}l}{\textit{Training}} \\
Batch size & 128 \\
Optimizer & Adam \\
Epochs & 1 \\
Learning rate & 2e-4 \\
LR scheduler & Cosine \\
Warmup ratio & 3\% \\
\midrule
\rowcolor{Gray}
\multicolumn{2}{@{}l}{\textit{Inference}} \\
Package & vLLM \\
Strategy & Greedy \\
\bottomrule
\end{tabular}%
}
\vspace{-0.25in}
\end{wraptable}

\paragraph{Implementation details.}

We evaluate \textbf{CheXagent-3B}~\citep{chen2024chexagent}, \textbf{MedGemma-4B}~\citep{sellergren2025medgemma}, and \textbf{Lingshu-7B}~\citep{xu2025lingshu}.
We first train baseline models without clinical context, generating reports directly from single image.
When training with C-SRRG, we use all available clinical context (\eg, indication, prior studies).
The only exception is CheXagent-3B on the C-SRRG-Impression, where we use only \emph{indication} due to training failure (detailed in \Cref{sec:limitation_in_chexagent-3b}).
We consider the two most recent prior studies with limited number of images (2/3/2 for CheXagent-3B/MedGemma-4B/Lingshu-7B) due to computational constraints.
We apply LoRA~\citep{hu2022lora} for fine-tuning, optimizing with Adam~\citep{kingma2014adam}, and use vLLM~\citep{kwon2023efficient} for inference.
Greedy decoding is adopted for \emph{reproducibility} consistent with benchmarking purpose.
All experiments run on a single \href{https://www.nvidia.com/en-us/data-center/h100}{NVIDIA H100} GPU.
Detailed hyperparameter settings are in \Cref{tab:hyperparameters}.

\textbf{Evaluation metrics.}
We use standard metrics, such as \textbf{BLEU}~\citep{papineni2002bleu}, \textbf{ROUGE-L}~\citep{lin2004rouge}, and \textbf{BERTScore}~\citep{zhang2019bertscore}, to assess text quality.
For clinical accuracy, we report \textbf{F1-RadGraph}~\citep{delbrouck2022improving} and SRRG-specific metrics~\citep{delbrouck2025automated}: \textbf{F1-SRRG-BERT}, built on CXR-BERT~\citep{boecking2022making} for structured evaluation, and \textbf{Category Score} (only for findings) for the correctness of organ-section headers.

\vspace{-0.05in}
\subsection{Experimental Results}
\label{sec:experimental_results}
\vspace{-5pt}




\begin{table}[ht]
\centering
\vspace{-0.05in}
\caption{\textbf{Results on the C-SRRG-Findings}. Clinical context is incorporated with our C-SRRG framework.}
\vspace{-0.1in}
\label{tab:findings_results}
\resizebox{\textwidth}{!}{%
\setlength{\tabcolsep}{3pt}
\begin{tabular}{lclcccccccccc}
\toprule
\multirow{3}{*}{\textbf{Model}} & \multirow{3}{*}{\textbf{\begin{tabular}{c}Clinical\\Context\end{tabular}}} & \multirow{3}{*}{\textbf{Split}} & \multicolumn{4}{c}{\textbf{Traditional Metrics}} & \multicolumn{3}{c}{\textbf{F1-SRR-BERT}} & \multicolumn{3}{c}{\textbf{Category Score}} \\
\cmidrule(lr){4-7} \cmidrule(lr){8-10} \cmidrule(lr){11-13}
& & & \textbf{BLEU} & \textbf{ROUGE-L} & \textbf{\begin{tabular}{c}BERT\\Score\end{tabular}} & \textbf{\begin{tabular}{c}F1-\\RadGraph\end{tabular}} & \textbf{Precision} & \textbf{Recall} & \textbf{\begin{tabular}{c}F1-\\Score\end{tabular}} & \textbf{Precision} & \textbf{Recall} & \textbf{\begin{tabular}{c}F1-\\Score\end{tabular}} \\
\midrule
\multirow{6}{*}{\textbf{CheXagent-3B}} &  & Valid & 1.97 & 20.63 & 30.33 & 13.07 & 44.67 & 45.16 & 43.46 & 73.61 & 81.17 & 75.54 \\
& \textcolor{Red}{\xmark} & Test & 2.08 & 20.09 & 31.91 & 12.99 & 43.73 & 42.54 & 41.70 & 74.47 & 85.26 & 77.74 \\
&  & Test-reviewed & 2.13 & 20.38 & 32.73 & 12.96 & 44.94 & 42.78 & 42.31 & 72.84 & 87.35 & 77.55 \\
& \cellcolor{Gray} & \cellcolor{Gray}Valid & \cellcolor{Gray}2.31 & \cellcolor{Gray}23.01 & \cellcolor{Gray}33.46 & \cellcolor{Gray}15.76 & \cellcolor{Gray}48.73 & \cellcolor{Gray}48.20 & \cellcolor{Gray}46.79 & \cellcolor{Gray}77.58 & \cellcolor{Gray}83.46 & \cellcolor{Gray}78.73 \\
& \cellcolor{Gray}\textcolor{Green}{\checkmark} & \cellcolor{Gray}Test & \cellcolor{Gray}1.89 & \cellcolor{Gray}20.92 & \cellcolor{Gray}33.28 & \cellcolor{Gray}13.58 & \cellcolor{Gray}45.18 & \cellcolor{Gray}44.10 & \cellcolor{Gray}43.07 & \cellcolor{Gray}75.79 & \cellcolor{Gray}85.69 & \cellcolor{Gray}78.82 \\
& \cellcolor{Gray} & \cellcolor{Gray}Test-reviewed & \cellcolor{Gray}1.98 & \cellcolor{Gray}21.64 & \cellcolor{Gray}34.32 & \cellcolor{Gray}14.05 & \cellcolor{Gray}47.50 & \cellcolor{Gray}45.09 & \cellcolor{Gray}44.59 & \cellcolor{Gray}76.08 & \cellcolor{Gray}88.87 & \cellcolor{Gray}79.93 \\
\midrule
\multirow{6}{*}{\textbf{MedGemma-4B}} &  & Valid & 1.51 & 20.95 & 30.83 & 13.98 & 42.93 & 45.50 & 42.12 & 78.48 & 78.00 & 76.26 \\
& \textcolor{Red}{\xmark} & Test & 1.58 & 19.69 & 31.52 & 13.30 & 42.32 & 41.38 & 40.19 & 76.31 & 82.36 & 77.44 \\
&  & Test-reviewed & 1.60 & 20.11 & 32.61 & 13.42 & 44.49 & 42.94 & 41.92 & 75.39 & 86.56 & 78.24 \\
& \cellcolor{Gray} & \cellcolor{Gray}Valid & \cellcolor{Gray}4.98 & \cellcolor{Gray}27.22 & \cellcolor{Gray}37.87 & \cellcolor{Gray}20.44 & \cellcolor{Gray}50.52 & \cellcolor{Gray}49.68 & \cellcolor{Gray}48.42 & \cellcolor{Gray}80.38 & \cellcolor{Gray}83.73 & \cellcolor{Gray}80.35 \\
& \cellcolor{Gray}\textcolor{Green}{\checkmark} & \cellcolor{Gray}Test & \cellcolor{Gray}3.05 & \cellcolor{Gray}23.17 & \cellcolor{Gray}35.65 & \cellcolor{Gray}15.91 & \cellcolor{Gray}45.84 & \cellcolor{Gray}44.24 & \cellcolor{Gray}43.43 & \cellcolor{Gray}78.28 & \cellcolor{Gray}84.67 & \cellcolor{Gray}79.59 \\
& \cellcolor{Gray} & \cellcolor{Gray}Test-reviewed & \cellcolor{Gray}4.29 & \cellcolor{Gray}24.37 & \cellcolor{Gray}36.60 & \cellcolor{Gray}17.01 & \cellcolor{Gray}47.90 & \cellcolor{Gray}45.17 & \cellcolor{Gray}44.96 & \cellcolor{Gray}76.73 & \cellcolor{Gray}87.84 & \cellcolor{Gray}80.04 \\
\midrule
\multirow{6}{*}{\textbf{Lingshu-7B}} &  & Valid & 1.42 & 17.68 & 27.20 & 10.56 & 40.15 & 41.45 & 39.29 & 74.37 & 75.97 & 73.57 \\
& \textcolor{Red}{\xmark} & Test & 1.40 & 17.71 & 29.65 & 11.14 & 40.60 & 39.41 & 38.65 & 75.86 & 81.47 & 76.84 \\
&  & Test-reviewed & 1.60 & 18.62 & 31.09 & 12.09 & 42.85 & 40.82 & 40.37 & 74.32 & 85.20 & 77.39 \\
& \cellcolor{Gray} & \cellcolor{Gray}Valid & \cellcolor{Gray}6.02 & \cellcolor{Gray}28.70 & \cellcolor{Gray}38.85 & \cellcolor{Gray}21.67 & \cellcolor{Gray}51.16 & \cellcolor{Gray}50.50 & \cellcolor{Gray}49.20 & \cellcolor{Gray}81.97 & \cellcolor{Gray}83.03 & \cellcolor{Gray}80.87 \\
& \cellcolor{Gray}\textcolor{Green}{\checkmark} & \cellcolor{Gray}Test & \cellcolor{Gray}3.16 & \cellcolor{Gray}23.53 & \cellcolor{Gray}35.60 & \cellcolor{Gray}16.07 & \cellcolor{Gray}45.96 & \cellcolor{Gray}44.42 & \cellcolor{Gray}43.63 & \cellcolor{Gray}79.80 & \cellcolor{Gray}83.20 & \cellcolor{Gray}79.68 \\
& \cellcolor{Gray} & \cellcolor{Gray}Test-reviewed & \cellcolor{Gray}4.42 & \cellcolor{Gray}23.70 & \cellcolor{Gray}35.76 & \cellcolor{Gray}16.09 & \cellcolor{Gray}47.48 & \cellcolor{Gray}44.80 & \cellcolor{Gray}44.54 & \cellcolor{Gray}77.57 & \cellcolor{Gray}86.71 & \cellcolor{Gray}79.83 \\
\bottomrule
\end{tabular}%
}
\vspace{-0.2in}
\end{table}
\paragraph{Results on the C-SRRG-Findings.}
\Cref{tab:findings_results} demonstrates \textbf{substantial improvements} achieved by C-SRRG on the C-SRRG-Findings across all evaluation metrics, except for slight BLEU decreases for CheXagent-3B on the test/test-reviewed splits (-0.19/-0.15).
For example, F1\text{-}SRR\text{-}BERT scores improve by \textbf{+3.33/+1.37/+2.28} (CheXagent-3B), \textbf{+6.30/+3.24/+3.04} (MedGemma-4B), and \textbf{+9.91/+4.98/+4.17} (Lingshu-7B), with \textbf{larger models consistently showing greater gains}. 
Category Score performance likewise improves by \textbf{+3.19/+0.99/+2.38} (CheXagent-3B), \textbf{+4.09/+2.15/+1.80} (MedGemma-4B), and \textbf{+7.30/+2.84/+2.44} (Lingshu-7B).


\vspace{-0.1in}
\paragraph{Results on the C-SRRG-Impression.}
\Cref{tab:impression_results} also shows \textbf{significant gains} achieved by our C-SRRG on the C-SRRG-Impression.
F1-SRR-BERT improves by \textbf{+0.8/+3.12} (CheXagent-3B), \textbf{+5.5/+4.43/+4.69} (MedGemma-4B), and \textbf{+7.42/+7.68/+6.16} (Lingshu-7B)
except for CheXagent-3B on the valid split (-0.06).
CheXagent-3B also exhibits similar BLEU score decreases on the C-SRRG-Findings  (\Cref{tab:findings_results}), indicating that rich clinical context may compromise text generation fluency in smaller models.
Importantly, while performance consistently drops as \textbf{models scale up without clinical context} from 3B to 7B parameters, it \textbf{improves substantially with context}, suggesting the \textbf{critical importance of clinical context} in scaling up MLLMs for SRRG.


\begin{table}[H]
\centering
\vspace{-0.25in}
\caption{\textbf{Results on the C-SRRG-Impression}. Clinical context is incorporated with our C-SRRG framework.}
\vspace{-0.13in}
\label{tab:impression_results}
\footnotesize
\begin{adjustbox}{width=\textwidth,center}
\begin{tabular}{lclccccccc}
\toprule
\multirow{3}{*}{\textbf{Model}} & \multirow{3}{*}{\textbf{\begin{tabular}{c}Clinical\\Context\end{tabular}}} & \multirow{3}{*}{\textbf{Split}} & \multicolumn{4}{c}{\textbf{Traditional Metrics}} & \multicolumn{3}{c}{\textbf{F1-SRR-BERT}} \\
\cmidrule(lr){4-7} \cmidrule(lr){8-10}
& & & \textbf{BLEU} & \textbf{ROUGE-L} & \textbf{\begin{tabular}{c}BERT\\Score\end{tabular}} & \textbf{\begin{tabular}{c}F1-\\RadGraph\end{tabular}} & \textbf{Precision} & \textbf{Recall} & \textbf{\begin{tabular}{c}F1-\\Score\end{tabular}} \\
\midrule
\multirow{6}{*}{\textbf{CheXagent-3B}} & & Valid & \phantom{0}9.44 & 34.03 & 61.82 & 19.30 & 63.80 & 63.48 & 59.10 \\
& \textcolor{Red}{\xmark} & Test & \phantom{0}7.83 & 29.40 & 59.82 & 16.13 & 57.18 & 59.18 & 54.27 \\
& & Test-reviewed & \phantom{0}7.42 & 28.60 & 58.35 & 13.71 & 51.32 & 56.34 & 49.74 \\
& \cellcolor{Gray} & \cellcolor{Gray}Valid & \cellcolor{Gray}\phantom{0}7.52 & \cellcolor{Gray}32.99 & \cellcolor{Gray}60.93 & \cellcolor{Gray}17.90 & \cellcolor{Gray}66.28 & \cellcolor{Gray}61.75 & \cellcolor{Gray}59.04 \\
& \cellcolor{Gray}\textcolor{Green}{\checkmark} & \cellcolor{Gray}Test & \cellcolor{Gray}\phantom{0}7.03 & \cellcolor{Gray}29.18 & \cellcolor{Gray}59.66 & \cellcolor{Gray}16.07 & \cellcolor{Gray}59.69 & \cellcolor{Gray}58.42 & \cellcolor{Gray}55.07 \\
& \cellcolor{Gray} & \cellcolor{Gray}Test-reviewed & \cellcolor{Gray}\phantom{0}6.92 & \cellcolor{Gray}29.04 & \cellcolor{Gray}58.91 & \cellcolor{Gray}14.84 & \cellcolor{Gray}55.42 & \cellcolor{Gray}58.26 & \cellcolor{Gray}52.86 \\
\midrule
\multirow{6}{*}{\textbf{MedGemma-4B}} & & Valid & \phantom{0}8.92 & 41.24 & 60.94 & 17.80 & 62.19 & 60.77 & 56.81 \\
& \textcolor{Red}{\xmark} & Test & \phantom{0}7.15 & 37.84 & 59.09 & 15.35 & 56.27 & 57.01 & 52.69 \\
& & Test-reviewed & \phantom{0}7.57 & 35.91 & 58.35 & 14.57 & 51.69 & 54.42 & 49.51 \\
& \cellcolor{Gray} & \cellcolor{Gray}Valid & \cellcolor{Gray}11.76 & \cellcolor{Gray}46.26 & \cellcolor{Gray}64.28 & \cellcolor{Gray}24.25 & \cellcolor{Gray}65.78 & \cellcolor{Gray}66.77 & \cellcolor{Gray}62.31 \\
& \cellcolor{Gray}\textcolor{Green}{\checkmark} & \cellcolor{Gray}Test & \cellcolor{Gray}10.58 & \cellcolor{Gray}41.92 & \cellcolor{Gray}61.85 & \cellcolor{Gray}19.23 & \cellcolor{Gray}59.45 & \cellcolor{Gray}61.89 & \cellcolor{Gray}57.12 \\
& \cellcolor{Gray} & \cellcolor{Gray}Test-reviewed & \cellcolor{Gray}11.21 & \cellcolor{Gray}40.15 & \cellcolor{Gray}61.12 & \cellcolor{Gray}19.16 & \cellcolor{Gray}55.02 & \cellcolor{Gray}60.71 & \cellcolor{Gray}54.20 \\
\midrule
\multirow{6}{*}{\textbf{Lingshu-7B}} & & Valid & \phantom{0}8.15 & 32.17 & 59.15 & 17.23 & 63.82 & 57.10 & 55.06 \\
& \textcolor{Red}{\xmark} & Test & \phantom{0}6.65 & 27.27 & 57.18 & 13.87 & 56.03 & 51.55 & 49.33 \\
& & Test-reviewed & \phantom{0}7.04 & 27.70 & 57.37 & 13.49 & 52.34 & 52.85 & 48.37 \\
& \cellcolor{Gray} & \cellcolor{Gray}Valid & \cellcolor{Gray}11.77 & \cellcolor{Gray}38.46 & \cellcolor{Gray}64.82 & \cellcolor{Gray}25.29 & \cellcolor{Gray}69.42 & \cellcolor{Gray}63.57 & \cellcolor{Gray}62.48 \\
& \cellcolor{Gray}\textcolor{Green}{\checkmark} & \cellcolor{Gray}Test & \cellcolor{Gray}10.58 & \cellcolor{Gray}32.86 & \cellcolor{Gray}62.07 & \cellcolor{Gray}19.85 & \cellcolor{Gray}63.04 & \cellcolor{Gray}58.39 & \cellcolor{Gray}57.01 \\
& \cellcolor{Gray} & \cellcolor{Gray}Test-reviewed & \cellcolor{Gray}11.61 & \cellcolor{Gray}33.66 & \cellcolor{Gray}62.04 & \cellcolor{Gray}21.28 & \cellcolor{Gray}57.48 & \cellcolor{Gray}58.80 & \cellcolor{Gray}54.53 \\
\bottomrule
\end{tabular}
\end{adjustbox}
\vspace{-0.2in}
\end{table}
\begin{table}[H]
\centering
\vspace{-0.1in}
\begin{minipage}[t]{0.48\textwidth}
\centering
\caption{\textbf{Effect of clinical context for train/eval on the C-SRR-Findings} using MedGemma-4B.}
\vspace{-0.13in}
\label{tab:findings_results_context}
\resizebox{\textwidth}{!}{%
\begin{tabular}{cclccc}
\toprule
\multicolumn{2}{c}{\textbf{Clinical Context}} & \multirow{2}{*}{\textbf{Split}} & \multicolumn{3}{c}{\textbf{F1-SRR-BERT}} \\
\cmidrule(lr){1-2} \cmidrule(lr){4-6}
\textbf{Train} & \textbf{Eval} & & \textbf{Precision} & \textbf{Recall} & \textbf{F1-Score} \\
\midrule
& & Valid & 42.93 & 45.50 & 42.12 \\
\textcolor{Red}{\xmark} & \textcolor{Red}{\xmark} & Test & 42.32 & 41.38 & 40.19 \\
& & Test-reviewed & 44.49 & 42.94 & 41.92 \\
\midrule
\rowcolor{Gray}
& & Valid & 47.00 & 47.09 & 45.35 \\
\rowcolor{Gray}
\textcolor{Green}{\checkmark} & \textcolor{Red}{\xmark} & Test & 42.76 & 41.55 & 40.64 \\
\rowcolor{Gray}
& & Test-reviewed & 44.79 & 43.12 & 42.45 \\
\midrule
& & Valid & 45.28 & 45.25 & 43.56 \\
\textcolor{Red}{\xmark} & \textcolor{Green}{\checkmark} & Test & 43.02 & 40.94 & 40.44 \\
& & Test-reviewed & 44.40 & 41.50 & 41.36 \\
\midrule
\rowcolor{Gray}
& & Valid & 50.52 & 49.68 & 48.42 \\
\rowcolor{Gray}
\textcolor{Green}{\checkmark} & \textcolor{Green}{\checkmark} & Test & 45.84 & 44.24 & 43.43 \\
\rowcolor{Gray}
& & Test-reviewed & 47.90 & 45.17 & 44.96 \\
\bottomrule
\end{tabular}%
}
\end{minipage}
\hfill
\begin{minipage}[t]{0.48\textwidth}
\centering
\caption{\textbf{Effect of clinical context for train/eval on the C-SRR-Impression} using MedGemma-4B.}
\vspace{-0.13in}
\label{tab:impression_results_context}
\resizebox{\textwidth}{!}{%
\begin{tabular}{cclccc}
\toprule
\multicolumn{2}{c}{\textbf{Clinical Context}} & \multirow{2}{*}{\textbf{Split}} & \multicolumn{3}{c}{\textbf{F1-SRR-BERT}} \\
\cmidrule(lr){1-2} \cmidrule(lr){4-6}
\textbf{Train} & \textbf{Eval} & & \textbf{Precision} & \textbf{Recall} & \textbf{F1-Score} \\
\midrule
& & Valid & 62.19 & 60.77 & 56.81 \\
\textcolor{Red}{\xmark} & \textcolor{Red}{\xmark} & Test & 56.27 & 57.01 & 52.69 \\
& & Test-reviewed & 51.69 & 54.42 & 49.51 \\
\midrule
\rowcolor{Gray}
& & Valid & 63.87 & 61.86 & 58.45 \\
\rowcolor{Gray}
\textcolor{Green}{\checkmark} & \textcolor{Red}{\xmark} & Test & 54.42 & 56.53 & 51.64 \\
\rowcolor{Gray}
& & Test-reviewed & 51.45 & 57.86 & 51.17 \\
\midrule
& & Valid & 62.60 & 64.23 & 59.10 \\
\textcolor{Red}{\xmark} & \textcolor{Green}{\checkmark} & Test & 53.34 & 59.11 & 52.59 \\
& & Test-reviewed & 49.35 & 58.68 & 50.66 \\
\midrule
\rowcolor{Gray}
& & Valid & 65.78 & 66.77 & 62.31 \\
\rowcolor{Gray}
\textcolor{Green}{\checkmark} & \textcolor{Green}{\checkmark} & Test & 59.45 & 61.89 & 57.12 \\
\rowcolor{Gray}
& & Test-reviewed & 55.02 & 60.71 & 54.20 \\
\bottomrule
\end{tabular}%
}
\end{minipage}
\vspace{-0.2in}
\end{table}
\begin{table}[H]
\centering
\vspace{-0.1in}
\begin{minipage}[t]{0.48\textwidth}
\centering
\caption{\textbf{Ablation study on clinical context for the C-SRRG-Findings} using MedGemma-4B.}
\vspace{-0.13in}
\label{tab:ablation_findings}
\resizebox{\textwidth}{!}{%
\begin{tabular}{llccc}
\toprule
\multicolumn{1}{c}{\textbf{Configuration}} & \multicolumn{1}{c}{\textbf{Split}} & \multicolumn{3}{c}{\textbf{F1-SRR-BERT}} \\
\cmidrule(lr){3-5}
& & \textbf{Precision} & \textbf{Recall} & \textbf{F1-Score} \\
\midrule
\multirow{3}{*}{Single-view} & \cellcolor{Gray}Valid & \cellcolor{Gray}47.00 & \cellcolor{Gray}47.09 & \cellcolor{Gray}45.35 \\
& \cellcolor{Gray}Test & \cellcolor{Gray}42.76 & \cellcolor{Gray}41.55 & \cellcolor{Gray}40.64 \\
& \cellcolor{Gray}Test-reviewed & \cellcolor{Gray}44.79 & \cellcolor{Gray}43.12 & \cellcolor{Gray}42.45 \\
\midrule
\multirow{3}{*}{Multi-view} & Valid & 47.46 & 47.21 & 45.80 \\
& Test & 44.44 & 42.57 & 41.92 \\
& Test-reviewed & 45.49 & 42.69 & 42.39 \\
\midrule
\multirow{3}{*}{+ Indication} & \cellcolor{Gray}Valid & \cellcolor{Gray}46.95 & \cellcolor{Gray}46.06 & \cellcolor{Gray}44.85 \\
& \cellcolor{Gray}Test & \cellcolor{Gray}44.62 & \cellcolor{Gray}42.56 & \cellcolor{Gray}41.98 \\
& \cellcolor{Gray}Test-reviewed & \cellcolor{Gray}45.92 & \cellcolor{Gray}43.14 & \cellcolor{Gray}42.79 \\
\midrule
\multirow{3}{*}{+ Technique} & Valid & 50.35 & 49.27 & 48.24 \\
& Test & 45.50 & 43.89 & 43.15 \\
& Test-reviewed & 47.48 & 44.60 & 44.42 \\
\midrule
\multirow{3}{*}{\makecell{+ Comparison \\ + Prior studies}} & \cellcolor{Gray}Valid & \cellcolor{Gray}50.52 & \cellcolor{Gray}49.68 & \cellcolor{Gray}48.42 \\
& \cellcolor{Gray}Test & \cellcolor{Gray}45.84 & \cellcolor{Gray}44.24 & \cellcolor{Gray}43.43 \\
& \cellcolor{Gray}Test-reviewed & \cellcolor{Gray}47.90 & \cellcolor{Gray}45.17 & \cellcolor{Gray}44.96 \\
\bottomrule
\end{tabular}%
}
\end{minipage}
\hfill
\begin{minipage}[t]{0.48\textwidth}
\centering
\caption{\textbf{Ablation study on clinical context for C-SRRG-Impression} using MedGemma-4B.}
\vspace{-0.13in}
\label{tab:ablation_impression}
\resizebox{\textwidth}{!}{%
\begin{tabular}{llccc}
\toprule
\multicolumn{1}{c}{\textbf{Configuration}} & \multicolumn{1}{c}{\textbf{Split}} & \multicolumn{3}{c}{\textbf{F1-SRR-BERT}} \\
\cmidrule(lr){3-5}
& & \textbf{Precision} & \textbf{Recall} & \textbf{F1-Score} \\
\midrule
\multirow{3}{*}{Single-view} & \cellcolor{Gray}Valid & \cellcolor{Gray}63.87 & \cellcolor{Gray}61.86 & \cellcolor{Gray}58.45 \\
& \cellcolor{Gray}Test & \cellcolor{Gray}54.42 & \cellcolor{Gray}56.53 & \cellcolor{Gray}51.64 \\
& \cellcolor{Gray}Test-reviewed & \cellcolor{Gray}51.45 & \cellcolor{Gray}57.86 & \cellcolor{Gray}51.17 \\
\midrule
\multirow{3}{*}{Multi-view} & Valid & 65.74 & 62.92 & 59.89 \\
& Test & 55.70 & 58.11 & 53.36 \\
& Test-reviewed & 51.78 & 59.37 & 52.25 \\
\midrule
\multirow{3}{*}{+ Indication} & \cellcolor{Gray}Valid & \cellcolor{Gray}66.91 & \cellcolor{Gray}65.02 & \cellcolor{Gray}61.67 \\
& \cellcolor{Gray}Test & \cellcolor{Gray}58.47 & \cellcolor{Gray}59.11 & \cellcolor{Gray}55.00 \\
& \cellcolor{Gray}Test-reviewed & \cellcolor{Gray}53.32 & \cellcolor{Gray}60.47 & \cellcolor{Gray}52.86 \\
\midrule
\multirow{3}{*}{+ Technique} & Valid & 65.91 & 65.65 & 61.78 \\
& Test & 58.65 & 60.24 & 55.88 \\
& Test-reviewed & 54.66 & 60.05 & 53.39 \\
\midrule
\multirow{3}{*}{\makecell{+ Comparison \\ + Prior studies}} & \cellcolor{Gray}Valid & \cellcolor{Gray}65.78 & \cellcolor{Gray}66.77 & \cellcolor{Gray}62.31 \\
& \cellcolor{Gray}Test & \cellcolor{Gray}59.45 & \cellcolor{Gray}61.89 & \cellcolor{Gray}57.12 \\
& \cellcolor{Gray}Test-reviewed & \cellcolor{Gray}55.02 & \cellcolor{Gray}60.71 & \cellcolor{Gray}54.20 \\
\bottomrule
\end{tabular}%
}
\end{minipage}
\vspace{-0.2in}
\end{table}

\vspace{-0.15in}
\paragraph{Effect of clinical context on training/evaluation.}
We next conduct ablation studies with four settings: 1) train+eval without context (baseline); 2) train with, eval without; 3) train without, eval with; and 4) train+eval with context.
\Cref{tab:findings_results_context,tab:impression_results_context} report F1-SRR-BERT on C-SRRG-Findings and C-SRRG-Impression using MedGemma-4B.
We find incorporating clinical context in only one phase provides limited improvement or slight degradation: \eg, +3.23/+0.45/+0.53 (train \textcolor{Green}{\checkmark}), or +1.44/+0.25/-0.56 (test \textcolor{Green}{\checkmark}) in findings and +1.65/-1.05/+1.66 (train \textcolor{Green}{\checkmark}), or +2.29/-0.1/+1.15 (test \textcolor{Green}{\checkmark}) in impression, which shows the \textbf{benefit of using context in both phases} for SRRG performance.


\vspace{-0.15in}
\paragraph{Impact of Each Clinical Context Component.}

We ablate four clinical-context components, 1) multi-view images, 2) indication, 3) technique, and 4) prior studies with comparison, to isolate their contributions on the performance.
\Cref{tab:ablation_findings,tab:ablation_impression} report F1-SRR-BERT for each variant on the C-SRRG-Findings, C-SRRG-Impression, respectively.
\textbf{All components contribute incrementally} to both tasks (except for few cases), with \textbf{performance being highest when using all available context}, which shows the importance of incorporating clinical context in SRRG.


\begin{wraptable}{r}{0.65\textwidth}
\centering
\caption{\textbf{Mitigation effect of temporal hallucination}.}
\label{tab:hallucination_results}
\vspace{-0.1in}
\small
\resizebox{\linewidth}{!}{%
\setlength{\tabcolsep}{4pt}
\begin{tabular}{@{}llccc@{}}
\toprule
\multirow{2}{*}{\textbf{Task}} & \multirow{2}{*}{\textbf{Split}} & \multicolumn{2}{c}{\textbf{Temporal Hallucination Rate}} & \multirow{2}{*}{\textbf{Mitigation}} \\
\cmidrule(lr){3-4}
& & \multicolumn{1}{c}{\textbf{Baseline (\textcolor{Red}{\xmark})}} & \multicolumn{1}{c}{\textbf{C-SRRG (\textcolor{Green}{\checkmark})}} & \\
\midrule
\multirow{4}{*}{Findings}
& Valid & \phantom{000}146/976\phantom{0} (15.0\%) & \phantom{000}70/976\phantom{00} (7.2\%) & \textcolor{blue}{-7.8\%} \\
& Test & \phantom{000}416/1459 (28.5\%) & \phantom{000}194/1459 (13.3\%) & \textcolor{blue}{-15.2\%} \\
& Test-reviewed & \phantom{0000}49/233\phantom{0} (21.0\%) & \phantom{0000}21/233\phantom{00} (9.0\%) & \textcolor{blue}{-12.0\%} \\
& \cellcolor{Gray}Overall & \cellcolor{Gray}\phantom{000}611/2668 (22.9\%) & \cellcolor{Gray}\phantom{000}285/2668 (10.7\%) & \cellcolor{Gray}\textcolor{blue}{-12.2\%} \\
\midrule
\multirow{4}{*}{Impression}
& Valid & \phantom{000}630/1505 (41.9\%) & \phantom{000}364/1505 (24.2\%) & \textcolor{blue}{-17.7\%} \\
& Test & \phantom{00}1012/2219 (45.6\%) & \phantom{000}599/2219 (27.0\%) & \textcolor{blue}{-18.6\%} \\
& Test-reviewed & \phantom{0000}92/231\phantom{0} (39.8\%) & \phantom{0000}58/231\phantom{0} (25.1\%) & \textcolor{blue}{-14.7\%} \\
& \cellcolor{Gray}Overall & \cellcolor{Gray}\phantom{00}1734/3955 (43.8\%) & \cellcolor{Gray}\phantom{00}1021/3955 (25.8\%) & \cellcolor{Gray}\textcolor{blue}{-18.0\%} \\
\bottomrule
\end{tabular}%
}
\vspace{-0.1in}
\end{wraptable}
\vspace{-0.1in}
\paragraph{Mitigation of temporal hallucinations.}

To quantify temporal hallucinations, we train MedGemma-4B under two conditions: \textbf{without} clinical context (\textbf{baseline}) and \textbf{with} clinical context (\textbf{C-SRRG}).
We evaluate both on evaluation sets \textbf{without clinical context}, and count reports that contain one of the following 33 indicators:
1) time references (`new', `newly', `recent', `recently', `previous', `prior', `interval', `compared to', `since', `from prior'),
2) stability indicators (`unchanged', `stable', `persistent', `persisting'), and
3) change indicators (`improved', `improvement', `worsened', `worsening', `increased', `decreased', `enlarging', `reducing', `progression', `regression', `evolving', `evolve', `developing', `developed', `resolving', `resolved', `temporal change', `compare', `comparison').
By this, we can detect whether the generated reports \textbf{contained hallucinations} by identifying inappropriate temporal references in the \textbf{absence of clinical context}. 
\Cref{tab:hallucination_results} shows that \textbf{clinical context substantially mitigates hallucinations}: Findings drop from 22.9\% to 10.7\% (–12.2\%) and Impression from 43.8\% to 25.8\% (–18.0\%).
This shows that C-SRRG effectively handles \emph{heterogeneous clinical context availability}, \ie, the absence of clinical context, while successfully mitigating temporal hallucinations.

\begin{wraptable}{r}{0.55\textwidth}
\centering
\vspace{-0.13in}
\caption{\textbf{Organ-level Category Score} on the Valid split.}
\vspace{-0.1in}
\label{tab:organ_level_comparison_validation}
\resizebox{\linewidth}{!}{%
\begin{tabular}{ccccccc}
\toprule
\multirow{2}{*}{\textbf{Region}} & \multicolumn{3}{c}{\textbf{Baseline (\textcolor{Red}{\xmark}})} & \multicolumn{3}{c}{\textbf{C-SRRG (\textcolor{Green}{\checkmark})}} \\
\cmidrule(lr){2-4} \cmidrule(lr){5-7}
& \textbf{Precision} & \textbf{Recall} & \textbf{F1-Score} & \textbf{Precision} & \textbf{Recall} & \textbf{F1-Score} \\
\midrule
P & 47.69 & 39.62 & 41.72 & 58.2 & 52.39 & \cellcolor{blue!20}52.77 \\
A & 8.0 & 8.0 & 8.0 & 17.86 & 17.86 & \cellcolor{blue!20}17.86 \\
H/M & 37.91 & 36.55 & \cellcolor{blue!20}36.97 & 34.25 & 33.07 & 33.41 \\
O & 9.21 & 7.94 & 8.2 & 12.17 & 10.13 & \cellcolor{blue!20}10.6 \\
L/A & 40.67 & 64.67 & 45.68 & 57.25 & 62.24 & \cellcolor{blue!20}55.69 \\
C & 71.77 & 68.6 & 68.99 & 73.28 & 70.26 & \cellcolor{blue!20}70.51 \\
M/C & 26.22 & 25.26 & 25.55 & 43.2 & 41.97 & \cellcolor{blue!20}42.3 \\
T/C/S & 57.04 & 64.18 & 58.45 & 59.81 & 64.39 & \cellcolor{blue!20}60.35 \\
\bottomrule
\end{tabular}%
}
\vspace{-0.1in}
\end{wraptable}
\vspace{-0.1in}
\paragraph{Anatomical region analysis.}
We compare organ-level performance for findings task against the baseline using the SRRG anatomical categories~\citep{delbrouck2025automated}.
\Cref{tab:organ_level_comparison_validation} reports the Category Score on the validation split using MedGemma-4B, with abbreviations: P = pleura, A = abdominal, H/M = hila/mediastinum, O = Other, L/A = lungs/airways, C = cardiovascular, M/C = musculoskeletal/chest wall, T/C/S = tubes/catheters/support devices.
We observe that incorporating clinical context with proposed C-SRRG \textbf{improves the performance across all the anatomical regions}, except for H/M.

\vspace{-0.25in}
\section{Conclusion, Limitations, and Future Work}
\label{sec:conclusion}
\vspace{-0.05in}

We introduced \emph{contextualized structured radiology report generation} (\textbf{C-SRRG}), a framework that aligns with radiologists' diagnostic workflow by integrating rich clinical context including multi-view images, indication, imaging technique, and prior studies with comparisons.
Through comprehensive evaluation of state-of-the-art medical MLLMs, we demonstrate that clinical context integration consistently enhances text quality, diagnostic accuracy, and reduces temporal hallucinations.
Importantly, our findings reveal that clinical context becomes \textbf{increasingly critical} as models scale up, suggesting that larger foundation models require more sophisticated contextual integration to achieve optimal performance.
We will publicly release our dataset, code, and model checkpoints to foster further research in C-SRRG and benefit the broader community.

\vspace{-0.13in}
\paragraph{Limitations.}



The C-SRRG dataset relies on synthetic LLM annotations from reformulated reports, which may introduce biases and subtle hallucinations.
Our supervised fine-tuning approach with greedy decoding may limit the full capture of clinical reasoning processes.
Computational and architectural constraints limited our evaluation to 7B parameter models with restricted multiple image processing capabilities (\eg, Lingshu-7B and CheXagent-3B limited to 2 images) and context windows that constrain comprehensive longitudinal history integration.
Additionally, our recency-based selection strategy for prioritizing recent studies, while capturing clinically relevant temporal information, may occasionally omit important historical context.

\textbf{Future work} should explore scaling to larger foundation models with extended-context capabilities, developing intelligent clinical context selection policies through learned strategies or retrieval-augmented approaches, and incorporating preference learning techniques with radiologists' feedback.
Expanding to comprehensive clinical modalities also presents promising avenues for enhanced diagnostic accuracy, with detailed discussions provided in \Cref{sec:risks_limitations}.

\section*{Reproducibility Statement}

We ensure reproducibility by building C-SRRG entirely from publicly available MIMIC-CXR~\citep{johnson2019mimiccxr} and CheXpert-Plus~\citep{chambon2024chexpert} datasets, with detailed documentation of our data processing pipeline including longitudinal patient history extraction and multi-view image integration in \Cref{sec:dataset}.
All experimental configurations are specified in \Cref{sec:experimental_setups}, including model hyperparameters (learning rate 2e-4, batch size 128, LoRA rank 32), exact data splits with patient-level separation, and standard evaluation metrics.
We commit to publicly releasing our complete codebase, the C-SRRG dataset with clinical context annotations, trained model checkpoints, and documentation for dataset recreation.
All experiments use reproducible libraries (Hugging Face PEFT~\citep{peft}, vLLM~\citep{kwon2023efficient}) on a single \href{https://www.nvidia.com/en-us/data-center/h100}{NVIDIA H100} GPU with fixed random seeds.

\section*{Ethics Statement}

Our work presents no new ethical concerns as C-SRRG is built entirely from existing de-identified public datasets (MIMIC-CXR and CheXpert Plus) that have undergone rigorous de-identification and received appropriate IRB approvals.
No additional patient data was collected for this work, and all privacy protections from the source datasets are maintained.
We acknowledge that automated report generation systems may produce hallucinations when referencing non-existent prior studies, which our work specifically addresses by incorporating comprehensive clinical context.
Our dataset and models are intended solely for research purposes and should not be used for clinical decision-making without appropriate validation and regulatory approval.
The computational requirements are modest (single GPU training), minimizing environmental impact while maintaining research accessibility.

\bibliographystyle{iclr2026_conference}
\bibliography{references}

\newpage
\appendix

\vspace{1.5em}
{\centering
\begin{tikzpicture}[remember picture, overlay=false]
\node[rectangle, rounded corners=12pt, fill=black!8, minimum width=10cm, minimum height=1.8cm] at (0.15,-0.15) {};
\node[rectangle, rounded corners=12pt, fill=black!15, minimum width=10cm, minimum height=1.8cm] (mainbox) at (0,0) {};
\node[rectangle, rounded corners=8pt, fill=white, minimum width=9.2cm, minimum height=1.4cm] (innerbox) at (0,0) {};
\draw[line width=3pt, color=black!60, rounded corners=2pt] (-4.8, -0.6) -- (-4.8, 0.6);
\draw[line width=3pt, color=black!60, rounded corners=2pt] (4.8, -0.6) -- (4.8, 0.6);
\fill[black!40] (-4.2, 0.5) circle (2pt);
\fill[black!40] (-4.2, -0.5) circle (2pt);
\fill[black!40] (4.2, 0.5) circle (2pt);
\fill[black!40] (4.2, -0.5) circle (2pt);
\node[font=\LARGE\bfseries\sffamily] at (0,0) {\textcolor{black!75}{APPENDIX}};
\draw[line width=1pt, color=black!30] (-2.5, -0.3) -- (2.5, -0.3);
\end{tikzpicture}
\par}
\vspace{1.5em}

This appendix provides comprehensive supplementary materials including limitations and future work discussions (\Cref{sec:risks_limitations}), specific limitations of CheXagent-3B with long clinical contexts (\Cref{sec:limitation_in_chexagent-3b}), detailed dataset statistics with patient-level data splits (\Cref{sec:dataset_stats}), prompt design examples for findings and impression generation (\Cref{sec:ift_prompt}), complete instruction fine-tuning examples with clinical context integration (\Cref{sec:ift_prompt_full}), and analysis of temporal hallucinations in radiology report generation (\Cref{sec:hallucination_analysis}).


\section{Limitations and Future Works}
\label{sec:risks_limitations}

\subsection{Synthetic Dataset and Annotations}

Our C-SRRG dataset builds upon the SRRG dataset~\citep{delbrouck2025automated}, which was generated through reformulation of free-form radiology reports using large language models.
This synthetic generation process introduces several potential limitations that warrant careful consideration.
The use of LLM-generated content may introduce subtle hallucinations or inconsistencies that could propagate through our training pipeline.
Additionally, the reformulation process may inadvertently introduce biases present in the underlying language models, potentially affecting the diversity and clinical accuracy of the generated reports.

\subsection{Model Architecture and Scale Limitations}

Our experimental evaluation faces several architectural constraints that limit the full potential of our approach.
First, we are constrained to backbone models with parameters up to 7B, which likely underestimates achievable performance.
The computational and memory requirements of larger models present practical limitations for comprehensive evaluation across multiple architectures and clinical contexts.

Second, the multimodal large language models employed were not originally designed to handle multiple images simultaneously.
In our experiments, we encountered specific limitations requiring tailored approaches.
\textbf{Lingshu-7B}~\citep{xu2025lingshu} was limited to 2 images due to computational efficiency constraints, while \textbf{CheXagent-3B}~\citep{chen2024chexagent} was similarly restricted to 2 images due to model constraints.
In contrast, \textbf{MedGemma-4B}~\citep{sellergren2025medgemma} demonstrated multi-image processing capabilities, requiring fewer tokens per image and enabling the use of more images in our longitudinal analysis setting.

These parameter and architectural constraints particularly impact the models' ability to effectively integrate complex temporal relationships and multi-modal clinical information across current and prior studies.
However, ongoing research trends in long-context LLM development~\citep{dao2022flashattention,kwon2023efficient} suggest that future model architectures will naturally address these limitations.
Scaling to larger foundation models, exploring mixture-of-experts variants, and advances in multimodal attention mechanisms represent promising directions for substantial improvements.

\subsection{Clinical Context Integration and Selection Limitations}

Our approach faces constraints in both the scope and selection of clinical contexts.
We impose limits on the number of images and prior studies included per clinical case, with prioritization given to the most recent studies.
This limitation stems from current multimodal architectures' context window constraints and computational overhead of processing extensive longitudinal histories.
While this recency-based selection strategy captures the most clinically relevant temporal information, it may occasionally omit important historical context that could inform diagnostic reasoning.
Additionally, our current implementation primarily utilizes publicly available datasets such as MIMIC-CXR~\citep{johnson2019mimiccxr} and CheXpert-Plus~\citep{chambon2024chexpert}, limiting the diversity of clinical contexts.

The scope could be significantly expanded to include additional clinical modalities such as CT imaging, Electronic Health Records (EHR)~\citep{hayrinen2008definition}, and comprehensive patient histories.
Future work should explore learned selection policies that intelligently identify the most informative clinical contexts and optimize longitudinal coverage.
Retrieval-augmented generation approaches over Picture Archiving and Communication Systems (PACS)~\citep{andriole2023picture} and EHR systems could dynamically surface the most relevant historical information for each case, unlocking the untapped potential for richer clinical context integration.

\subsection{Training Methodology and Decoding Limitations}

Our training approach is restricted to supervised fine-tuning with greedy decoding for reproducibility and computational efficiency.
This methodology, while providing stable and consistent results, may not fully capture the nuanced decision-making processes that characterize expert radiological interpretation.
The supervised learning paradigm limits the model's ability to learn from comparative feedback and iterative refinement that occurs in clinical practice.
Incorporating preference learning techniques and Reinforcement Learning (RL)-based methods with radiologists' feedback, such as Proximal Policy Optimization (PPO)~\citep{schulman2017proximal} or Direct Preference Optimization (DPO)~\citep{rafailov2023direct}, could enhance the fidelity and clinical appropriateness of generated reports.
Furthermore, exploring retrieval-conditioned decoding strategies could improve temporal consistency and reduce hallucinations by grounding generation in verified clinical contexts.


\section{Limitation in CheXagent-3B on C-SRRG-Impression}
\label{sec:limitation_in_chexagent-3b}

While most models show improvement with clinical context, CheXagent-3B exhibits a \textbf{critical failure} in following the structured report format instructions when provided with full clinical context.
Instead of generating properly formatted impression sections with numbered findings, the model frequently produces single-word outputs or generic phrases.
For instance, when the expected format is a multi-point structured impression such as ``1. Slight decrease in size of the right apicolateral pneumothorax with chest tube in place. 2. Unchanged multifocal right-sided pulmonary opacities...'', CheXagent-3B often generates only ``Pneumothorax'' or ``Pneumonia''.
This format degradation is widespread, with the model generating non-structured outputs like ``No acute cardiopulmonary process'' or ``Pulmonary edema'' rather than detailed clinical impressions.

\begin{table}[ht]
\centering
\caption{\textbf{Performance degradation of CheXagent-3B on C-SRRG-Impression with full clinical context.} The model shows dramatic drops across all metrics when provided with complete clinical context.}
\label{tab:chexagent_context_failure}
\small
\begin{adjustbox}{width=\textwidth,center}
\begin{tabular}{lclccccccc}
\toprule
\multirow{3}{*}{\textbf{Model}} & \multirow{3}{*}{\textbf{\begin{tabular}{c}Full\\Clinical\\Context\end{tabular}}} & \multirow{3}{*}{\textbf{Split}} & \multicolumn{4}{c}{\textbf{Traditional Metrics}} & \multicolumn{3}{c}{\textbf{F1-SRR-BERT}} \\
\cmidrule(lr){4-7} \cmidrule(lr){8-10}
& & & \textbf{BLEU} & \textbf{ROUGE-L} & \textbf{\begin{tabular}{c}BERT\\Score\end{tabular}} & \textbf{\begin{tabular}{c}F1-\\RadGraph\end{tabular}} & \textbf{Precision} & \textbf{Recall} & \textbf{\begin{tabular}{c}F1-\\Score\end{tabular}} \\
\midrule
\multirow{6}{*}{\textbf{CheXagent-3B}} & & Valid & 9.44 & 34.03 & 61.82 & 19.30 & 63.80 & 63.48 & 59.10 \\
& \textcolor{Red}{\xmark} & Test & 7.83 & 29.40 & 59.82 & 16.13 & 57.18 & 59.18 & 54.27 \\
& & Test-reviewed & 7.42 & 28.60 & 58.35 & 13.71 & 51.32 & 56.34 & 49.74 \\
& \cellcolor{Gray} & \cellcolor{Gray}Valid & \cellcolor{Gray}2.57 & \cellcolor{Gray}21.76 & \cellcolor{Gray}40.10 & \cellcolor{Gray}13.10 & \cellcolor{Gray}74.48 & \cellcolor{Gray}49.40 & \cellcolor{Gray}54.05 \\
& \cellcolor{Gray}\textcolor{Green}{\checkmark} & \cellcolor{Gray}Test & \cellcolor{Gray}2.40 & \cellcolor{Gray}17.54 & \cellcolor{Gray}33.79 & \cellcolor{Gray}\phantom{0}9.78 & \cellcolor{Gray}66.56 & \cellcolor{Gray}41.04 & \cellcolor{Gray}45.99 \\
& \cellcolor{Gray} & \cellcolor{Gray}Test-reviewed & \cellcolor{Gray}2.89 & \cellcolor{Gray}19.61 & \cellcolor{Gray}37.88 & \cellcolor{Gray}11.87 & \cellcolor{Gray}64.18 & \cellcolor{Gray}41.27 & \cellcolor{Gray}46.44 \\
\bottomrule
\end{tabular}
\end{adjustbox}
\end{table}

The performance metrics in \Cref{tab:chexagent_context_failure} reveal the severity of this issue: when provided with full clinical context, traditional metrics plummet dramatically (BLEU: 9.44→2.57, ROUGE-L: 34.03→21.76, BERTScore: 61.82→40.10 on validation set).
This catastrophic degradation suggests that CheXagent-3B, likely trained primarily on shorter sequence lengths, struggles to process and integrate the extensive clinical context while maintaining adherence to the structured output format.
The model's inability to handle long input sequences effectively undermines its utility for clinical applications requiring comprehensive context integration.



\section{Detailed Dataset Statistics}
\label{sec:dataset_stats}

We provide detailed statistics of our clinical context chest X-ray dataset, focusing on patient distribution across splits.

\paragraph{Patient Distribution Across Splits.}

Our dataset maintains strict patient-level separation across training, validation, and test splits to prevent data leakage.
As shown in the patient overlap heatmaps, the training set contains 83,147 unique patients for findings and 125,947 unique patients for impression tasks.
The validation sets include 434 patients for findings and 477 patients for impression, while the test sets contain 274 patients for findings and 423 patients for impression.
The test-reviewed splits comprise 173 patients for findings and 172 patients for impression, with 106 and 108 patients respectively shared with the test split.
This patient-level split ensures that clinical studies from the same patient do not appear across different evaluation splits, with zero patient overlap between training and evaluation sets.
The distribution maintains clinical diversity while preserving the integrity of comprehensive clinical contexts within patient histories.

\begin{figure}[htbp]
    \centering
    \begin{subfigure}[b]{0.48\textwidth}
        \centering
        \includegraphics[width=\textwidth]{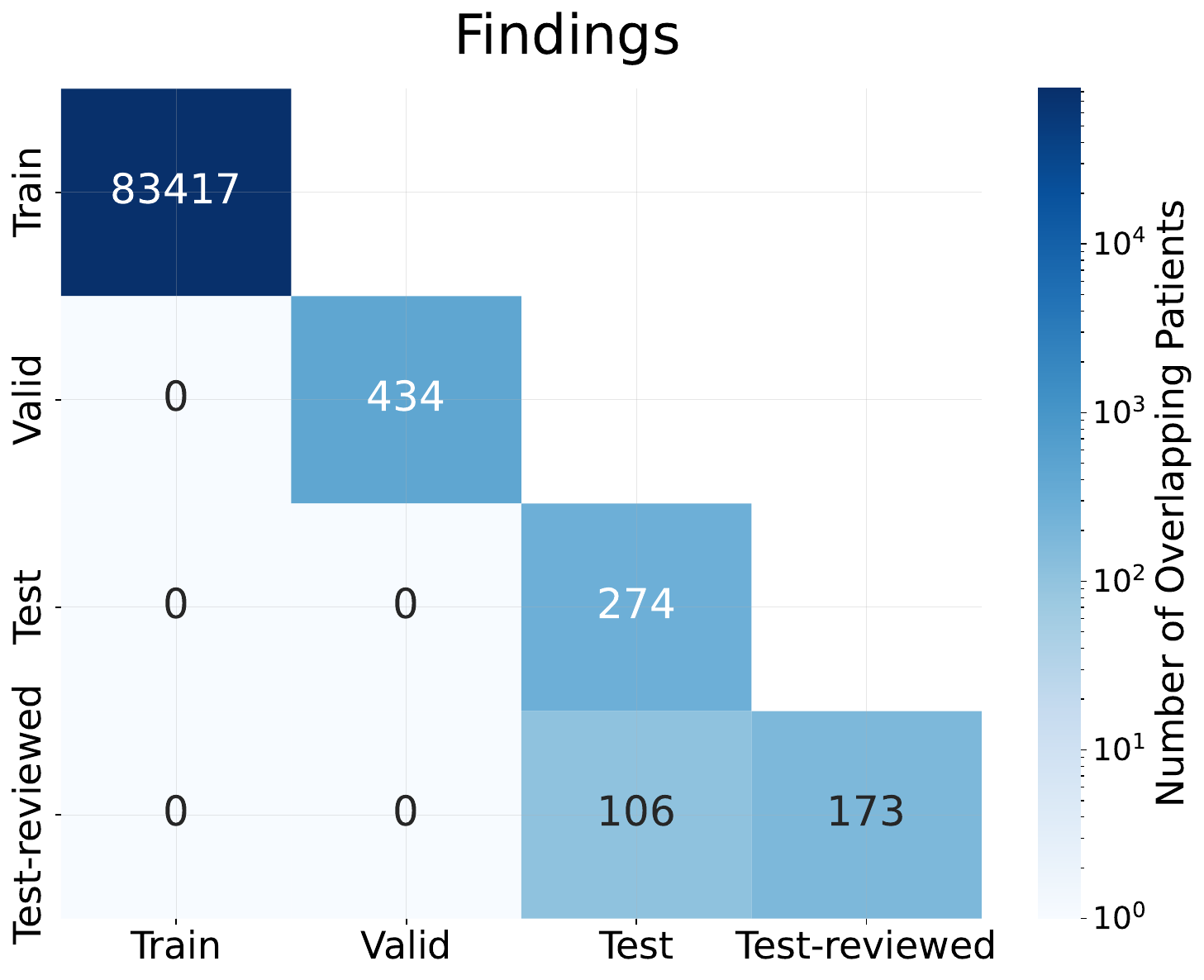}
        \label{fig:patient_overlap_findings}
    \end{subfigure}
    \quad
    \begin{subfigure}[b]{0.48\textwidth}
        \centering
        \includegraphics[width=\textwidth]{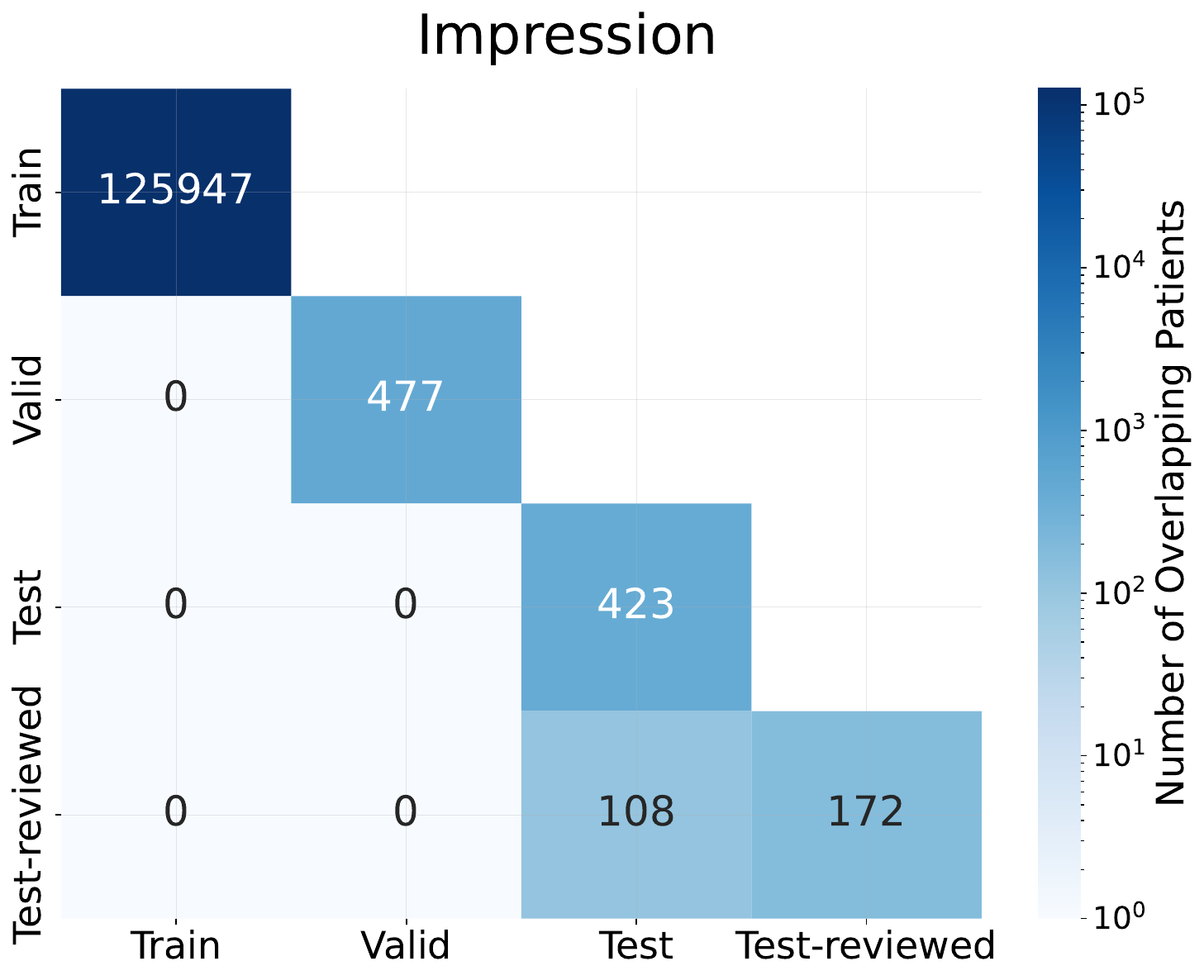}
        \label{fig:patient_overlap_impression}
    \end{subfigure}
    \vspace{-0.2in}
    \caption{\textbf{Patient overlap heatmaps} across train, valid, test, and test-reviewed splits.}
    \label{fig:patient_overlap_heatmaps}
\end{figure}



\section{Prompt Designs}
\label{sec:ift_prompt}

In this section, beyond formats for impression with prior studies (\Cref{fig:prompt_format_impression}), current study (\Cref{fig:current_study_format}), we provide examples of our design choices for prompts, \eg, formats for findings with/without prior studies (\Cref{fig:prompt_format_finding,fig:prompt_format_finding_no_history}), impression without prior studies (\Cref{fig:prompt_format_impression_no_history}), prior studies (\Cref{fig:prior_study_format}), and response format for findings and impression (\Cref{fig:response_format_findings,fig:response_format_impression}), used for training and evaluation.

\begin{figure}[!tb]
    \centering
    \vspace{-0.1in}
    \begin{tcolorbox}[
        colback=gray!5,
        colframe=gray!40,
        width=\textwidth,
        boxrule=0.8pt,
        arc=2mm,
        left=3mm,
        right=3mm,
        top=2mm,
        bottom=2mm
    ]
    \small
    \textbf{SYSTEM PROMPT:}\\
    You are an expert radiologist.
    \\[1.5em]
    \textbf{USER PROMPT:}\\
    Analyze the current chest X-ray images and compare them with the previous studies to write the FINDINGS section of a radiology report.
    Use standard medical terminology and note any changes from the prior studies, focusing on the most recent comparisons.
    Consider the available clinical contexts when formulating your findings.
    \end{tcolorbox}
    \vspace{-0.2in}
    \caption{\textbf{Prompt format for FINDINGS} for C-SRRG.}
    \label{fig:prompt_format_finding}
\end{figure}

\begin{figure}[!tb]
    \centering
    \vspace{-0.1in}
    \begin{tcolorbox}[
        colback=gray!5,
        colframe=gray!40,
        width=\textwidth,
        boxrule=0.8pt,
        arc=2mm,
        left=3mm,
        right=3mm,
        top=2mm,
        bottom=2mm
    ]
    \small
    \textbf{SYSTEM PROMPT:}\\
    You are an expert radiologist.
    \\[1.5em]
    \textbf{USER PROMPT:}\\
    Analyze the chest X-ray images and write the FINDINGS section of a radiology report.
    Use standard medical terminology and organize findings by anatomical regions.
    Consider the available clinical contexts when formulating your findings.
    \end{tcolorbox}
    \vspace{-0.2in}
    \caption{\textbf{Prompt format for FINDINGS without previous history} for C-SRRG.}
    \label{fig:prompt_format_finding_no_history}
\end{figure}

\begin{figure}[!tb]
    \centering
    \vspace{-0.1in}
    \begin{tcolorbox}[
        colback=gray!5,
        colframe=gray!40,
        width=\textwidth,
        boxrule=0.8pt,
        arc=2mm,
        left=3mm,
        right=3mm,
        top=2mm,
        bottom=2mm
    ]
    \small
    \textbf{SYSTEM PROMPT:}\\
    You are an expert radiologist.
    \\[1.5em]
    \textbf{USER PROMPT:}\\
    Analyze the chest X-ray images and write the IMPRESSION section of a radiology report.
    Provide a concise clinical summary and diagnosis based on the imaging findings.
    Consider the available clinical contexts when formulating your impression.
    \end{tcolorbox}
    \vspace{-0.2in}
    \caption{\textbf{Prompt format for IMPRESSION without previous history} for C-SRRG.}
    \label{fig:prompt_format_impression_no_history}
\end{figure}

\begin{figure}[!tb]
    \centering
    \vspace{-0.1in}
    \begin{tcolorbox}[
        colback=gray!5,
        colframe=gray!40,
        width=\textwidth,
        boxrule=0.8pt,
        arc=2mm,
        left=3mm,
        right=3mm,
        top=2mm,
        bottom=2mm
    ]
    \small
    \textbf{=== PREVIOUS STUDY N ===}
    
    Previous Clinical History/Indication:\\
    Status post coronary artery bypass grafting (CABG), post chest tube removal.
    \\[0.5em]
    Previous Technique:\\
    Standard posteroanterior and lateral chest radiographs.
    \\[0.5em]
    Previous Findings:\\
    Lungs and Airways:\\
    - No pneumothorax visualized\\
    - Low lung volumes\\
    - Increased infiltrate in the left upper lung
    \\[0.5em]
    Pleura:\\
    - Retrocardiac opacity\\
    - Layering left pleural effusion
    \\[0.5em]
    Cardiovascular:\\
    - Vascular plethora
    \\[0.5em]
    Other:\\
    - Patchy areas of alveolar edema on the right\\
    - Endotracheal (ET) tube positioned 4.3 cm above the carina\\
    - Nasogastric (NG) tube within the stomach
    \\[0.5em]
    Previous Impression:\\
    1. Worsened infiltrate in the left upper lung suggestive of an acute process.\\
    2. Presence of retrocardiac opacity and layering left pleural effusion.\\
    3. Vascular plethora and patchy alveolar edema on the right, which may indicate fluid overload or cardiac etiology.\\
    4. Proper positioning of the ET and NG tubes.
    \\[0.5em]
    Previous Images:\\
    \texttt{[Images: Previous chest X-ray images]}
    \end{tcolorbox}
    \vspace{-0.2in}
    \caption{\textbf{An example of prior study} for C-SRRG.}
    \label{fig:prior_study_format}
\end{figure}

\begin{figure}[!tb]
    \vspace{-0.1in}
    \centering
    \begin{tcolorbox}[
        colback=gray!5,
        colframe=gray!40,
        width=\textwidth,
        boxrule=0.8pt,
        arc=2mm,
        left=3mm,
        right=3mm,
        top=2mm,
        bottom=2mm
    ]
    \small
    \textbf{ASSISTANT RESPONSE:}\\
    FINDINGS:\\
    Lungs and Airways:\\
    - No pneumothorax or focal consolidation.
    \\[0.5em]
    Pleura:\\
    - Enlarged small right pleural effusion.
    \\[0.5em]
    Cardiovascular:\\
    - Mildly enlarged heart.
    \\[0.5em]
    Hila and Mediastinum:\\
    - Normal hilar and mediastinal contours.
    \\[0.5em]
    Tubes, Catheters, and Support Devices:\\
    - Right internal jugular (IJ) catheter terminates at the superior cavoatrial junction.
    \end{tcolorbox}
    \vspace{-0.2in}
    \caption{\textbf{An example of ground-truth assistant responses} in the C-SRRG-Findings dataset.}
    \label{fig:response_format_findings}
    \vspace{-0.1in}
\end{figure}



\section{Instruction Tuning Dataset Prompt Example}
\label{sec:ift_prompt_full}

We provide detailed instruction fine-tuning examples that showcase the comprehensive clinical context utilized in our approach.
These examples demonstrate how all available clinical information is systematically integrated into our instruction tuning dataset, including patient medical history, imaging techniques, previous study findings, and temporal comparisons.
The following multi-part examples illustrate the complete structure of our training data, highlighting how comprehensive clinical contexts including temporal, multi-view, and metadata information are preserved and leveraged for clinical reasoning in radiology report generation.

\subsection{Findings Generation Example}
\label{sec:findings_example}

The first example demonstrates the generation of the FINDINGS section, which requires detailed anatomical observation and temporal comparison across multiple studies (\Cref{fig:findings_example_part1,fig:findings_example_part2,fig:findings_example_part3}):

\begin{figure}[!tb]
\begin{tcolorbox}[
    colback=gray!5,
    colframe=gray!40,
    width=\textwidth,
    boxrule=0.8pt,
    arc=2mm,
    left=3mm,
    right=3mm,
    top=2mm,
    bottom=2mm,
    title=Findings Example - Part 1: Current Study Context,
    fonttitle=\bfseries
]
\small
\textbf{USER PROMPT:}\\
Analyze the current chest X-ray images and compare them with the previous studies to write the FINDINGS section of a radiology report.
Use standard medical terminology and note any changes from the prior studies, focusing on the most recent comparisons.
Consider the available clinical contexts when formulating your findings.
\\[1.0em]
\textbf{=== CURRENT CLINICAL HISTORY/INDICATION ===}\\
Evaluation for fluid overload.
\\[1.0em]
\textbf{=== CURRENT TECHNIQUE ===}\\
Standard frontal chest radiography protocol.
\\[1.0em]
\textbf{=== CURRENT COMPARISON ===}\\
Prior radiographs and CT scans.
\\[1.0em]
\textbf{=== CURRENT IMAGES ===}\\
\texttt{[Images: Current chest X-ray images]}
\end{tcolorbox}
\vspace{-0.2in}
\caption{\textbf{Findings generation example (Part 1)} in the C-SRRG-Findings dataset.}
\label{fig:findings_example_part1}
\end{figure}

\begin{figure}[!tb]
\begin{tcolorbox}[
    colback=gray!5,
    colframe=gray!40,
    width=\textwidth,
    boxrule=0.8pt,
    arc=2mm,
    left=3mm,
    right=3mm,
    top=2mm,
    bottom=2mm,
    title=Findings Example - Part 2: Previous Study 1,
    fonttitle=\bfseries
]
\small
\textbf{=== PREVIOUS STUDY 1 (Most Recent) ===}

Previous Clinical History/Indication:\\
Patient with a history of multifocal after CABG, currently presenting with symptoms suggestive of CHF or pneumonia.
\\[0.5em]
Previous Technique:\\
A single frontal chest radiograph was obtained.
\\[0.5em]
Previous Comparison:\\
Multiple prior radiographs
\\[0.5em]
Previous Findings:\\
Lungs and Airways:\\
- No definitive consolidation observed on this examination; however, subsequent CT confirms presence at the right base\\
- Mild pulmonary edema
\\[0.5em]
Pleura:\\
- Moderate right pleural effusion, unchanged\\
- No pneumothorax
\\[0.5em]
Cardiovascular:\\
- Moderate cardiomegaly noted\\
- Aortic tortuosity present
\\[0.5em]
Tubes, Catheters, and Support Devices:\\
- Status post median sternotomy with CABG and valve replacements
\\[0.5em]
Previous Impression:\\
1. Mild pulmonary edema with right pleural effusion.\\
2. Right base consolidation not definitive on radiograph but confirmed on subsequent CT.
\\[0.5em]
Previous Images:\\
\texttt{[Images: Previous chest X-ray images]}
\end{tcolorbox}
\vspace{-0.2in}
\caption{\textbf{Findings generation example (Part 2)} in the C-SRRG-Findings dataset.}
\label{fig:findings_example_part2}
\end{figure}

\begin{figure}[!tb]
\begin{tcolorbox}[
    colback=gray!5,
    colframe=gray!40,
    width=\textwidth,
    boxrule=0.8pt,
    arc=2mm,
    left=3mm,
    right=3mm,
    top=2mm,
    bottom=2mm,
    title=Findings Example - Part 3: Previous Study 2 \& Response,
    fonttitle=\bfseries
]
\small
\textbf{=== PREVIOUS STUDY 2 ===}

Previous Clinical History/Indication:\\
The patient presents with tachycardia.
\\[0.5em]
Previous Technique:\\
Posteroanterior (PA) and lateral chest radiographs.
\\[0.5em]
Previous Findings:\\
Cardiovascular:\\
- Status post sternotomy with mitral and aortic valve replacements\\
- Moderate cardiomegaly, unchanged
\\[0.5em]
Hila and Mediastinum:\\
- Stable mediastinal and hilar contours
\\[0.5em]
Lungs and Airways:\\
- New mild-to-moderate elevation of the right hemidiaphragm suggesting volume loss\\
- Increased patchy opacity in the right lower lobe, concerning for pneumonic consolidation\\
- Probable new opacification in the right middle lobe\\
- New vague opacity in the right suprahilar region of the right upper lobe, potentially indicating early pneumonia
\\[0.5em]
Pleura:\\
- No definite pleural effusion
\\[0.5em]
Previous Impression:\\
1. Findings suggest multifocal pneumonia involving the right lung.
\\[1.0em]
\textbf{ASSISTANT RESPONSE:}\\
FINDINGS:\\
Lungs and Airways:\\
- No pneumothorax or focal consolidation.
\\[0.5em]
Pleura:\\
- Enlarged small right pleural effusion.
\\[0.5em]
Cardiovascular:\\
- Mildly enlarged heart.
\\[0.5em]
Hila and Mediastinum:\\
- Normal hilar and mediastinal contours.
\\[0.5em]
Tubes, Catheters, and Support Devices:\\
- Right internal jugular (IJ) catheter terminates at the superior cavoatrial junction.
\end{tcolorbox}
\vspace{-0.2in}
\caption{\textbf{Findings generation example (Part 3)} in the C-SRRG-Findings dataset.}
\label{fig:findings_example_part3}
\end{figure}

\subsection{Impression Generation Example}
\label{sec:impression_example}

The second example demonstrates the generation of the IMPRESSION section, which requires clinical synthesis and diagnostic reasoning (\Cref{fig:impression_example_part1,fig:impression_example_part2,fig:impression_example_part3}):

\begin{figure}[!tb]
\begin{tcolorbox}[
    colback=gray!5,
    colframe=gray!40,
    width=\textwidth,
    boxrule=0.8pt,
    arc=2mm,
    left=3mm,
    right=3mm,
    top=2mm,
    bottom=2mm,
    title=Example Case with Clinical Context - Part 1: Current Study,
    fonttitle=\bfseries
]
\small
\textbf{USER PROMPT:}\\
Analyze the current chest X-ray images and compare them with the previous studies to write the IMPRESSION section of a radiology report.
Provide a concise clinical summary and diagnosis, noting any changes from the prior studies, focusing on the most recent comparisons.
Consider the available clinical contexts when formulating your impression.
\\[1.0em]
\textbf{=== CURRENT CLINICAL HISTORY/INDICATION ===}\\
Male with end-stage renal disease on hemodialysis, multiple orthopedic hardware, bioprosthetic aortic valve replacement, coronary artery disease status post coronary artery bypass grafting, permanent pacemaker for sick sinus syndrome, admitted for MRSA bacteremia, transferred to the critical care unit for hypotension during anesthesia induction.
Patient intubated for respiratory status assessment.
\\[1.0em]
\textbf{=== CURRENT TECHNIQUE ===}\\
Chest single view
\\[1.0em]
\textbf{=== CURRENT COMPARISON ===}\\
Prior imaging at an unspecified time.
\\[1.0em]
\textbf{=== CURRENT IMAGES ===}\\
\texttt{[Images: Current chest X-ray images]}
\end{tcolorbox}
\vspace{-0.2in}
\caption{\textbf{Impression generation example (Part 1)} in the C-SRRG-Impression dataset.}
\label{fig:impression_example_part1}
\end{figure}

\begin{figure}[!tb]
\begin{tcolorbox}[
    colback=gray!5,
    colframe=gray!40,
    width=\textwidth,
    boxrule=0.8pt,
    arc=2mm,
    left=3mm,
    right=3mm,
    top=2mm,
    bottom=2mm,
    title=Example Case with Clinical Context - Part 2: Previous Study 1,
    fonttitle=\bfseries
]
\small
\textbf{=== PREVIOUS STUDY 1 (Most Recent) ===}

Previous Clinical History/Indication:\\
Status post coronary artery bypass grafting (CABG), post chest tube removal.
\\[0.5em]
Previous Technique:\\
Standard posteroanterior and lateral chest radiographs.
\\[0.5em]
Previous Findings:\\
Lungs and Airways:\\
- No pneumothorax visualized\\
- Low lung volumes\\
- Increased infiltrate in the left upper lung
\\[0.5em]
Pleura:\\
- Retrocardiac opacity\\
- Layering left pleural effusion
\\[0.5em]
Cardiovascular:\\
- Vascular plethora
\\[0.5em]
Other:\\
- Patchy areas of alveolar edema on the right\\
- Endotracheal (ET) tube positioned 4.3 cm above the carina\\
- Nasogastric (NG) tube within the stomach
\\[0.5em]
Previous Impression:\\
1. Worsened infiltrate in the left upper lung suggestive of an acute process.\\
2. Presence of retrocardiac opacity and layering left pleural effusion.\\
3. Vascular plethora and patchy alveolar edema on the right, which may indicate fluid overload or cardiac etiology.\\
4. Proper positioning of the ET and NG tubes.
\\[0.5em]
Previous Images:\\
\texttt{[Images: Previous chest X-ray images]}
\end{tcolorbox}
\vspace{-0.2in}
\caption{\textbf{Impression generation example (Part 2)} in the C-SRRG-Impression dataset.}
\label{fig:impression_example_part2}
\end{figure}

\begin{figure}[!tb]
\begin{tcolorbox}[
    colback=gray!5,
    colframe=gray!40,
    width=\textwidth,
    boxrule=0.8pt,
    arc=2mm,
    left=3mm,
    right=3mm,
    top=2mm,
    bottom=2mm,
    title=Example Case with Clinical Context - Part 3: Previous Study 2 \& Response,
    fonttitle=\bfseries
]
\small
\textbf{=== PREVIOUS STUDY 2 ===}

Previous Clinical History/Indication:\\
Patient presents with portions and exertion x.
\\[0.5em]
Previous Technique:\\
The examination was performed using frontal and lateral views of the chest.
\\[0.5em]
Previous Findings:\\
Tubes, Catheters, and Support Devices:\\
- Dual lead left-sided pacemaker extending to the expected positions of the right atrium and right ventricle.
\\[0.5em]
Pleura:\\
- Slight blunting of the posterior costophrenic angles, which may suggest very trace pleural effusions.
\\[0.5em]
Lungs and Airways:\\
- No focal consolidation observed.
\\[0.5em]
Cardiovascular:\\
- The cardiac silhouette appears stable.
\\[0.5em]
Hila and Mediastinum:\\
- The mediastinal silhouette is stable.
\\[0.5em]
Musculoskeletal and Chest Wall:\\
- Right proximal humerus hardware is present but not well evaluated due to the imaging technique.
\\[0.5em]
Other:\\
- Slight prominence of the interstitium, which may indicate minimal interstitial edema.
\\[0.5em]
Previous Impression:\\
1. Possible very trace pleural effusions.\\
2. Minimal interstitial edema.\\
3. Stable cardiac and mediastinal silhouettes.\\
4. Presence of dual lead left-sided pacemaker in expected positions.\\
5. Right proximal humerus hardware present but not adequately assessed.
\\[1.0em]
\textbf{ASSISTANT RESPONSE:}\\
IMPRESSION:\\
1. Worsened left basilar opacity, which may represent atelectasis or pneumonitis in the appropriate clinical context.\\
2. Mild improvement in pulmonary vascularity.
\end{tcolorbox}
\vspace{-0.2in}
\caption{\textbf{Impression generation example (Part 3)} in the C-SRRG-Impression dataset.}
\label{fig:impression_example_part3}
\end{figure}


\section{Hallucination Analysis}
\label{sec:hallucination_analysis}

In this section, we examine a critical limitation of radiology report generation models trained without clinical context, specifically their tendency to hallucinate temporal comparisons when referencing non-existent prior studies.
We first demonstrate that dataset ground truth reports contain temporal statements that become hallucinations when clinical context is absent, as radiologists naturally write these comparisons when they have access to prior studies.
We then analyze how models trained without such clinical context systematically produce these hallucinations, even for patients with no imaging history.
Finally, we quantify these hallucinations by detecting the frequency of temporal statements on the generated reports on the evaluation set without clinical context.

\paragraph{Dataset Hallucination.}

Ground truth radiology reports in clinical datasets frequently contain temporal statements such as ``new from prior exam,'' ``unchanged,'' or ``stable compared to previous study.''
These temporal references are clinically appropriate when radiologists have access to prior imaging studies for comparison.
However, when language models are trained on these reports without access to the corresponding clinical context and prior studies, they learn to replicate these temporal language patterns indiscriminately.
This training paradigm creates a systematic hallucination problem where models generate temporal comparison statements even for patients with no prior imaging history.
The following examples demonstrate these temporal hallucinations present in ground truth radiology reports from the dataset, showing how temporal comparison statements appear without proper clinical context (\Cref{fig:hallucination_example1,fig:hallucination_example2,fig:hallucination_example3}):

\begin{figure}[!tb]
\begin{tcolorbox}[
    colback=red!5,
    colframe=red!40,
    width=\textwidth,
    boxrule=0.8pt,
    arc=2mm,
    left=3mm,
    right=3mm,
    top=2mm,
    bottom=2mm,
    title=Dataset Hallucination Example 1: Temporal Information Fabrication,
    fonttitle=\bfseries
]
\small
\textbf{Structured Report:}\\
Exam Type: Chest radiograph.
\\[0.5em]
Technique: Portable anteroposterior (AP) chest radiography was performed.
\\[0.5em]
History: A male patient with hep C cirrhosis and large right pleural effusion status post thoracocentesis.
Evaluate for resolution of pleural effusion.
\\[0.5em]
Comparison: Prior portable AP chest radiograph
\\[0.5em]
Findings:
\\[0.5em]
Lungs and Airways:\\
- Mild inflation of the right upper lobe\\
- Collapsed right lower lobe\\
- No consolidation in the left lung
\\[0.5em]
Pleura:\\
- Moderate pleural effusion within the right pleural space.\\
- \colorbox{red!20}{Moderate right pneumothorax, new from prior exam.}\\
- No left pleural effusion or pneumothorax.
\\[0.5em]
Cardiovascular:\\
- No significant mediastinal shift observed.
\\[0.5em]
Hila and Mediastinum:\\
- Mediastinum appears unremarkable
\\[0.5em]
Impression:\\
1. Moderate right-sided pneumothorax.\\
2. Moderate right pleural effusion.\\
3. Inflation of the right upper lobe with collapse of the right lower lobe.\\
4. No mediastinal shift.
\begin{tcolorbox}[
    colback=yellow!10,
    colframe=orange!60,
    boxrule=1pt,
    arc=1mm,
    left=2mm,
    right=2mm,
    top=1mm,
    bottom=1mm
]
\textbf{Hallucination:} The phrase ``\textbf{new from prior exam}'' represents temporal information that cannot be verified from the current study alone, if not with previous history.
\end{tcolorbox}
\end{tcolorbox}
\vspace{-0.2in}
\caption{\textbf{Dataset hallucination example 1} in SRRG dataset.}
\label{fig:hallucination_example1}
\end{figure}

\begin{figure}[!tb]
\begin{tcolorbox}[
    colback=red!5,
    colframe=red!40,
    width=\textwidth,
    boxrule=0.8pt,
    arc=2mm,
    left=3mm,
    right=3mm,
    top=2mm,
    bottom=2mm,
    title=Dataset Hallucination Example 2: Stability Assumption Without Comparison,
    fonttitle=\bfseries
]
\small
\textbf{Structured Report:}\\
Exam Type: Chest radiograph
\\[0.5em]
Technique: Standard frontal and lateral chest radiographic views were performed.
\\[0.5em]
History: Atrial fibrillation (AF), coronary artery disease (CAD), congestive heart failure (CHF).
\\[0.5em]
Comparison: Prior chest radiographs
\\[0.5em]
Findings:
\\[0.5em]
Cardiovascular:\\
- \colorbox{red!20}{Mild to moderate cardiomegaly, unchanged.}\\
- Tortuous but stable aorta.
\\[0.5em]
Tubes, Catheters, and Support Devices:\\
- Transvenous pacemaker/AICD with leads terminating in the right atrium and right ventricle.\\
- Median sternotomy wires are aligned and intact.
\\[0.5em]
Lungs and Airways:\\
- Lungs are clear with no evidence of consolidation, pleural effusion, pneumothorax, or overt pulmonary edema.
\\[0.5em]
Impression:\\
1. No radiographic evidence for acute cardiopulmonary process.
\begin{tcolorbox}[
    colback=yellow!10,
    colframe=orange!60,
    boxrule=1pt,
    arc=1mm,
    left=2mm,
    right=2mm,
    top=1mm,
    bottom=1mm
]
\textbf{Hallucination:} The term ``\textbf{unchanged}'' implies comparison with prior studies, which is problematic without previous history.
\end{tcolorbox}
\end{tcolorbox}
\vspace{-0.2in}
\caption{\textbf{Dataset hallucination example 2} in SRRG dataset.}
\label{fig:hallucination_example2}
\end{figure}

\begin{figure}[!tb]
\begin{tcolorbox}[
    colback=red!5,
    colframe=red!40,
    width=\textwidth,
    boxrule=0.8pt,
    arc=2mm,
    left=3mm,
    right=3mm,
    top=2mm,
    bottom=2mm,
    title=Dataset Hallucination Example 3: Generic Temporal Statement,
    fonttitle=\bfseries
]
\small
\textbf{Structured Report:}\\
Exam Type: Chest radiograph
\\[0.5em]
Technique: Single AP upright portable chest radiograph.
\\[0.5em]
History: Shortness of breath.
\\[0.5em]
Comparison: Prior chest radiograph
\\[0.5em]
Findings:
\\[0.5em]
Lungs and Airways:\\
- Low lung volumes\\
- Minimal bibasilar atelectasis\\
- Subcentimeter left lower lung rounded calcification, stable, possibly representing a calcified granuloma\\
- No focal consolidation\\
- No overt pulmonary edema
\\[0.5em]
Pleura:\\
- No pleural effusion\\
- No evidence of pneumothorax
\\[0.5em]
Cardiovascular:\\
- Stable cardiac and mediastinal silhouettes
\\[0.5em]
Hila and Mediastinum:\\
- Ovoid calcification over the left mediastinum, stable
\\[0.5em]
Tubes, Catheters, and Support Devices:\\
- Right-sided Port-A-Cath in place, terminating at the cavoatrial junction, without significant change in position
\\[0.5em]
Musculoskeletal and Chest Wall:\\
- Status post median sternotomy
\\[0.5em]
Impression:\\
1. \colorbox{red!20}{No significant interval change.}
\begin{tcolorbox}[
    colback=yellow!10,
    colframe=orange!60,
    boxrule=1pt,
    arc=1mm,
    left=2mm,
    right=2mm,
    top=1mm,
    bottom=1mm
]
\textbf{Hallucination:} The phrase ``\textbf{No significant interval change}'' is a generic temporal statement that, if not with prior studies, represents a non-specific conclusion that could mask important clinical details.
\end{tcolorbox}
\end{tcolorbox}
\vspace{-0.2in}
\caption{\textbf{Dataset hallucination example 3} in SRRG dataset.}
\label{fig:hallucination_example3}
\end{figure}

\paragraph{Trained Model Hallucination.}

Our trained models also exhibit hallucination patterns, particularly when generating temporal comparisons despite having access only to the current study.
The following examples from MedGemma-4B model trained without clinical history demonstrate how models can inappropriately generate comparative language even when no prior studies are available for comparison (\Cref{fig:trained_hallucination_example1,fig:trained_hallucination_example2,fig:trained_hallucination_example3,fig:trained_hallucination_example4,fig:trained_hallucination_example5,fig:trained_hallucination_example6}):

\begin{figure}[!tb]
\begin{tcolorbox}[
    colback=teal!5,
    colframe=teal!40,
    width=\textwidth,
    boxrule=0.8pt,
    arc=2mm,
    left=3mm,
    right=3mm,
    top=2mm,
    bottom=2mm,
    title=Trained Model Hallucination Example 1: Temporal Comparisons in Findings,
    fonttitle=\bfseries
]
\textbf{Sample ID:} 12

\textbf{Generated Report:}\\
FINDINGS:\\
Tubes, Catheters, and Support Devices:\\
- Endotracheal tube terminates 4.5 cm above the carina.\\
- Right internal jugular central venous catheter terminates in the right atrium.\\
- Left pectoral pacemaker with leads in the right atrium and right ventricle.\\
- Median sternotomy wires are intact.\\

Cardiovascular:\\
- \colorbox{red!20}{Moderate cardiomegaly, unchanged.}\\

Lungs and Airways:\\
- Low lung volumes.\\
- No evidence of pneumothorax.\\

Pleura:\\
- \colorbox{red!20}{Small left pleural effusion, unchanged.}\\

Other:\\
- \colorbox{red!20}{Retrocardiac opacity, likely representing atelectasis, unchanged.}

\begin{tcolorbox}[
    colback=yellow!10,
    colframe=orange!60,
    boxrule=1pt,
    arc=1mm,
    left=2mm,
    right=2mm,
    top=1mm,
    bottom=1mm
]
\textbf{Hallucination:} Multiple instances of ``\textbf{unchanged}'' indicate temporal comparisons when only analyzing a single study.
\end{tcolorbox}
\end{tcolorbox}
\vspace{-0.2in}
\caption{\textbf{Trained model hallucination example 1 on test-reviewed split} using MedGemma-4B.}
\label{fig:trained_hallucination_example1}
\end{figure}

\begin{figure}[!tb]
\begin{tcolorbox}[
    colback=teal!5,
    colframe=teal!40,
    width=\textwidth,
    boxrule=0.8pt,
    arc=2mm,
    left=3mm,
    right=3mm,
    top=2mm,
    bottom=2mm,
    title=Trained Model Hallucination Example 2: Interval and Persistent Language,
    fonttitle=\bfseries
]
\textbf{Sample ID:} 60

\textbf{Generated Report:}\\
FINDINGS:\\
Pleura:\\
- \colorbox{red!20}{Interval placement of a right-sided pigtail catheter.}\\
- \colorbox{red!20}{Decreased size of the right pleural effusion.}\\
- No evidence of pneumothorax.\\

Lungs and Airways:\\
- \colorbox{red!20}{Persistent right basilar opacity}, likely representing atelectasis.\\
- Left lung is clear.\\

Cardiovascular:\\
- Stable cardiomediastinal silhouette.

\begin{tcolorbox}[
    colback=yellow!10,
    colframe=orange!60,
    boxrule=1pt,
    arc=1mm,
    left=2mm,
    right=2mm,
    top=1mm,
    bottom=1mm
]
\textbf{Hallucination:} Terms like ``\textbf{Interval placement}'', ``\textbf{Decreased size}'', and ``\textbf{Persistent}'' all imply temporal comparisons.
\end{tcolorbox}
\end{tcolorbox}
\vspace{-0.2in}
\caption{\textbf{Trained model hallucination example 2 on test-reviewed split} using MedGemma-4B.}
\label{fig:trained_hallucination_example2}
\end{figure}

\begin{figure}[!tb]
\begin{tcolorbox}[
    colback=teal!5,
    colframe=teal!40,
    width=\textwidth,
    boxrule=0.8pt,
    arc=2mm,
    left=3mm,
    right=3mm,
    top=2mm,
    bottom=2mm,
    title=Trained Model Hallucination Example 3: Explicit Temporal Language,
    fonttitle=\bfseries
]
\textbf{Sample ID:} 37

\textbf{Generated Report:}\\
FINDINGS:\\
Lungs and Airways:\\
- \colorbox{red!20}{No significant interval change in the appearance of the lungs.}\\

Pleura:\\
- Small right pleural effusion, \colorbox{red!20}{unchanged.}\\

Cardiovascular:\\
- Stable cardiomediastinal silhouette.\\

Tubes, Catheters, and Support Devices:\\
- Tracheostomy tube in place.\\
- Right internal jugular (IJ) dialysis catheter in place.\\
- Vascular stent in the right subclavian region.\\

Other:\\
- No evidence of pneumothorax.

\begin{tcolorbox}[
    colback=yellow!10,
    colframe=orange!60,
    boxrule=1pt,
    arc=1mm,
    left=2mm,
    right=2mm,
    top=1mm,
    bottom=1mm
]
\textbf{Hallucination:} Explicit temporal language (``\textbf{No significant interval change}'', ``\textbf{unchanged}'') when no prior studies are available.
\end{tcolorbox}
\end{tcolorbox}
\vspace{-0.2in}
\caption{\textbf{Trained model hallucination example 3 on test-reviewed split} using MedGemma-4B.}
\label{fig:trained_hallucination_example3}
\end{figure}

\begin{figure}[!tb]
\begin{tcolorbox}[
    colback=teal!5,
    colframe=teal!40,
    width=\textwidth,
    boxrule=0.8pt,
    arc=2mm,
    left=3mm,
    right=3mm,
    top=2mm,
    bottom=2mm,
    title=Trained Model Hallucination Example 4: Unchanged Position in Impression,
    fonttitle=\bfseries
]
\textbf{Sample ID:} 0

\textbf{Generated Report:}\\
IMPRESSION:\\
1. No evidence of pneumothorax.\\
2. Stable small bilateral pleural effusions.\\
3. \colorbox{red!20}{Unchanged position of the right chest tube.}\\
4. Normal cardiomediastinal silhouette.\\
5. Presence of a vagal nerve stimulator.

\begin{tcolorbox}[
    colback=yellow!10,
    colframe=orange!60,
    boxrule=1pt,
    arc=1mm,
    left=2mm,
    right=2mm,
    top=1mm,
    bottom=1mm
]
\textbf{Hallucination:} The term ``\textbf{Unchanged position}'' suggests comparison with prior imaging that is not available.
\end{tcolorbox}
\end{tcolorbox}
\vspace{-0.2in}
\caption{\textbf{Trained model hallucination example 4 on test-reviewed split} using MedGemma-4B.}
\label{fig:trained_hallucination_example4}
\end{figure}

\begin{figure}[!tb]
\begin{tcolorbox}[
    colback=teal!5,
    colframe=teal!40,
    width=\textwidth,
    boxrule=0.8pt,
    arc=2mm,
    left=3mm,
    right=3mm,
    top=2mm,
    bottom=2mm,
    title=Trained Model Hallucination Example 5: Interval and Persistent Terms,
    fonttitle=\bfseries
]
\textbf{Sample ID:} 36

\textbf{Generated Report:}\\
IMPRESSION:\\
1. \colorbox{red!20}{Interval increase in bilateral pleural effusions.}\\
2. \colorbox{red!20}{Persistent left lower lobe atelectasis.}\\
3. Mild pulmonary edema.\\
4. Stable cardiomegaly.

\begin{tcolorbox}[
    colback=yellow!10,
    colframe=orange!60,
    boxrule=1pt,
    arc=1mm,
    left=2mm,
    right=2mm,
    top=1mm,
    bottom=1mm
]
\textbf{Hallucination:} Terms ``\textbf{Interval increase}'' and ``\textbf{Persistent}'' indicate temporal comparisons without access to prior studies.
\end{tcolorbox}
\end{tcolorbox}
\vspace{-0.2in}
\caption{\textbf{Trained model hallucination example 5 on test-reviewed split} using MedGemma-4B.}
\label{fig:trained_hallucination_example5}
\end{figure}

\begin{figure}[!tb]
\begin{tcolorbox}[
    colback=teal!5,
    colframe=teal!40,
    width=\textwidth,
    boxrule=0.8pt,
    arc=2mm,
    left=3mm,
    right=3mm,
    top=2mm,
    bottom=2mm,
    title=Trained Model Hallucination Example 6: New vs Unchanged Terminology,
    fonttitle=\bfseries
]
\textbf{Sample ID:} 40

\textbf{Generated Report:}\\
IMPRESSION:\\
1. \colorbox{red!20}{New right lower lobe opacity}, which may represent pneumonia or aspiration.\\
2. Small right pleural effusion.\\
3. Stable mild cardiomegaly.\\
4. \colorbox{red!20}{Unchanged position of the right-sided central venous catheter.}

\begin{tcolorbox}[
    colback=yellow!10,
    colframe=orange!60,
    boxrule=1pt,
    arc=1mm,
    left=2mm,
    right=2mm,
    top=1mm,
    bottom=1mm
]
\textbf{Hallucination:} The term ``\textbf{Unchanged position}'' implies knowledge of prior imaging studies that the model does not have access to.
\textbf{New} also suggests temporal comparison but may be acceptable if referring to acute findings.
\end{tcolorbox}
\end{tcolorbox}
\vspace{-0.2in}
\caption{\textbf{Trained model hallucination example 6 on test-reviewed split} using MedGemma-4B.}
\label{fig:trained_hallucination_example6}
\end{figure}

\clearpage
\section{The Use of LLMs}\label{sec:llm}
We used LLMs solely for light editing such as correcting grammatical errors and polishing some words. They did not contribute to research ideation, experiments, analysis, or substantive writing.

\end{document}